\documentclass[10pt,twocolumn,letterpaper]{article}

\usepackage[pagenumbers]{cvpr} 

\usepackage{graphicx}
\usepackage{amsmath}
\usepackage{amssymb}
\usepackage{booktabs}
\usepackage{multirow}
\usepackage{multicol}
\usepackage[american]{babel}
\usepackage{microtype}
\usepackage{booktabs,bigstrut}
\graphicspath{{Figure/}}

\usepackage{color, xcolor, colortbl}
\definecolor{citecolor}{HTML}{229954}
\usepackage[pagebackref=true,breaklinks=true,colorlinks,citecolor=citecolor,bookmarks=false]{hyperref}

\newcommand{\bfsection}[1]{\vspace*{0.1cm}\noindent\textbf{#1.}}

\usepackage[capitalize]{cleveref}
\crefname{section}{Sec.}{Secs.}
\Crefname{section}{Section}{Sections}
\Crefname{table}{Table}{Tables}
\crefname{table}{Tab.}{Tabs.}

\def\cvprPaperID{3101} 
\def\confName{CVPR}
\def\confYear{2022}

\begin{document}
	
	\title{Discrete Cosine Transform Network for Guided Depth Map Super-Resolution}
		
	\author{
		Zixiang Zhao$^{1,2}$\quad 
		Jiangshe Zhang$^{1}$\thanks{Corresponding authors.} \quad
		Shuang Xu$^{1,3*}$\quad
		Zudi Lin$^{2}$\quad
		Hanspeter Pfister$^{2}$\\[1mm]
		$^{1}$Xi’an Jiaotong University, Xi’an, China\\
		$^{2}$Harvard University, Cambridge MA, USA\\
		$^{3}$Northwestern Polytechnical University, Xi’an, China\\
		{\tt\small zixiangzhao@stu.xjtu.edu.cn, jszhang@mail.xjtu.edu.cn, xs@nwpu.edu.cn,}\\
		{\tt\small \{linzudi,pfister\}@g.harvard.edu}
	}
	\maketitle
	
	\begin{abstract}
    	Guided depth super-resolution~(GDSR) is an essential topic in multi-modal image processing, which reconstructs high-resolution (HR) depth maps from low-resolution ones collected with suboptimal conditions with the help of HR RGB images of the same scene. To solve the challenges in interpreting the working mechanism, extracting cross-modal features and RGB texture over-transferred, we propose a novel Discrete Cosine Transform Network~(DCTNet) to alleviate the problems from three aspects. First, the Discrete Cosine Transform~(DCT) module reconstructs the multi-channel HR depth features by using DCT to solve the channel-wise optimization problem derived from the image domain. Second, we introduce a semi-coupled feature extraction module that uses shared convolutional kernels to extract common information and private kernels to extract modality-specific information. Third, we employ an edge attention mechanism to highlight the contours informative for guided upsampling. Extensive quantitative and qualitative evaluations demonstrate the effectiveness of our DCTNet, which outperforms previous state-of-the-art methods with a relatively small number of parameters.
    	The code is available at \url{https://github.com/Zhaozixiang1228/GDSR-DCTNet}.
	\end{abstract}
	
\section{Introduction}\label{sec:1}
	
With the popularity of consumer-oriented depth estimation sensors, \eg, Time-of-Flight~(ToF) and Kinect cameras, depth maps have promoted advancements in autonomous driving \cite{DBLP:conf/eccv/LiaoLZZLY20,DBLP:conf/cvpr/PengPLS20}, pose estimation \cite{DBLP:conf/iccv/XiongZ0CYZY19,DBLP:journals/pami/ShottonGFSCFMKCKB13}, virtual reality \cite{DBLP:journals/tip/LiuZCJZG19a,DBLP:journals/tii/LiLY0J021}, and scene understanding \cite{DBLP:conf/iccv/ZhangBKIX17,DBLP:conf/eccv/GuptaGAM14}.
Unfortunately, due to the technical limitations and suboptimal imaging conditions, depth images are often low-resolution~(LR) and noisy. 
However, high-resolution~(HR) RGB images~(or intensity images) are relatively easy to obtain in the same scene when acquiring depth maps.
Therefore, {\em guided depth map super-resolution} (GDSR) with RGB images has become an essential topic in multi-modal image processing and multi-modal super-resolution~(SR).
Our research is based on the assumption that there are statistical co-occurrences between the texture edges of RGB images and the discontinuities of depth maps~\cite{DBLP:conf/eccv/RieglerRB16}.
In this way, information in RGB images can be utilized to restore HR depth maps when the LR depth maps are unsatisfactory for downstream applications.

\begin{figure}[t]
	\centering
	\includegraphics[width=\linewidth]{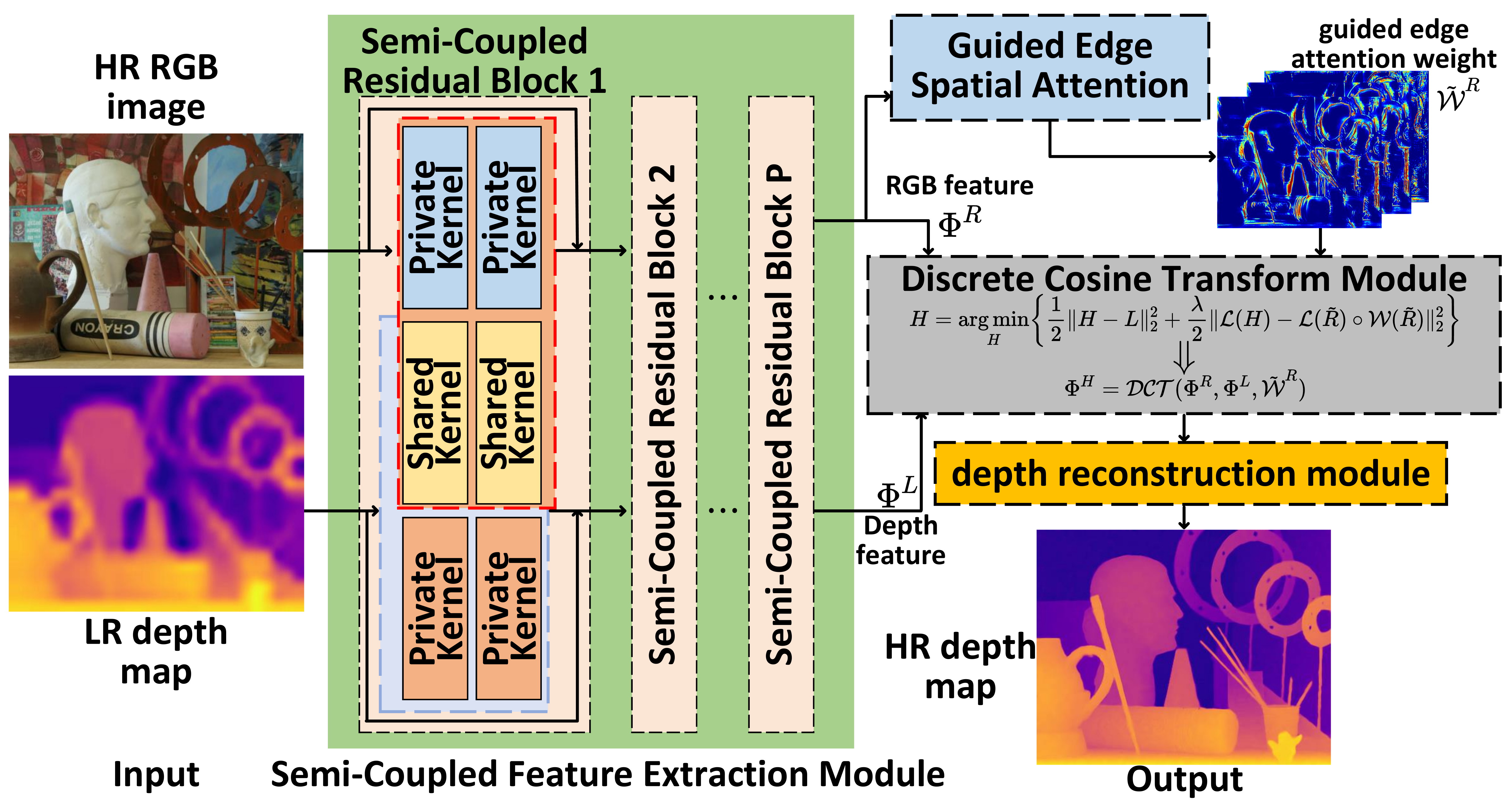}
	\vspace{-1.8em}
	\caption{
    Overview of DCTNet. First, the SCFE module extracts shared and private features from the depth (LR) and RGB (HR) images. The GESA module employs the RGB feature to obtain edge attention weights useful for SR. The multi-modal features and attention weights are then processed by the DCT module, where DCT is utilized in each channel to get HR depth features. Finally, the reconstruction module outputs the SR depth map.
	}
	\label{fig:Workflow}
\end{figure}

For image SR, deep neural networks have become the {\em de facto} methodology due to their ability in modeling the mapping from LR to HR images~\cite{DBLP:conf/eccv/DongLHT14,lim2017enhanced,DBLP:conf/cvpr/ZhangGT20}.
However, image SR mainly focuses on reconstructing fine details and textures, while depth SR models need to infer textureless and piecewise affine regions that have sharp depth discontinuities~\cite{DBLP:conf/eccv/RieglerRB16}. Besides, depth maps can be noisy and suffer from a lower tolerance for artifacts in real-world applications~\cite{DBLP:journals/tip/XieFS16}.
Therefore, we can hardly adopt the methods for image SR without appraising the unique characteristics of depth SR.

Conventional methods for GDSR can be divided into three categories, \ie, filter-~\cite{DBLP:conf/cvpr/0001TT13,DBLP:journals/tip/MinLD12,DBLP:conf/cvpr/LuSMLD12,DBLP:conf/cvpr/LuF15}, optimization-~\cite{DBLP:conf/nips/DiebelT05,DBLP:conf/iccv/ParkKTBK11,DBLP:journals/tip/YangYLHW14,DBLP:conf/iccv/FerstlRRRB13,DBLP:conf/eccv/LiMDL16} and learning-based~\cite{DBLP:journals/tmm/XieFYS15,DBLP:conf/cvpr/GuZGCCZ17,DBLP:journals/tip/XieFS16} methods. 
Filter-based (or local) methods focus on preserving sharp depth edges under the guidance of the intensity image. However, for texture-rich RGB images, irrelevant edges may be transferred to depth images (known as texture {\em over-transferred}). In addition, the explicitly defined filters can only model a specific visual task and lack flexibility.
Optimization-based (or global) methods design energy functions based on diverse data prior, with data-fidelity regularization terms constraint the solution space~\cite{DBLP:conf/cvpr/ZhangZGZ17,DBLP:journals/corr/abs-2109-00212}. 
However, natural priors are often challenging to be explicitly represented and learned.
The third category contains learning-based methods, which employ data-driven pipelines to learn the dependency between multi-modal inputs. Representative works in this category use sparse dictionary learning~\cite{DBLP:conf/iccv/KiechleHK13,DBLP:journals/tip/TosicD14,DBLP:conf/cvpr/KwonTL15}, which learn dictionaries in a group learning manner and set constraints on the sparse representations of different modalities~\cite{DBLP:conf/cvpr/ZhangZGZ17,DBLP:journals/tip/DengD20}.

Deep learning (DL) models are introduced to learn the mapping from LR to HR images~\cite{DBLP:journals/tip/YeSWYXLL20,DBLP:journals/tip/WenSLLF19,DBLP:conf/cvpr/SongDZLLLY20,DBLP:conf/cvpr/SunYLLW021,DBLP:conf/mm/TangCSHZZK21,DBLP:conf/mm/TangCZ21}, but they still often cooperate classic methods for depth upsampling. For example, learnable filter~\cite{DBLP:conf/cvpr/Wu0ZH18,DBLP:journals/ijcv/KimPH21} (combination of DL and filter-based methods) and algorithm unrolling~\cite{DBLP:journals/pami/0002D21,DBLP:conf/cvpr/Xu0ZSL021} (DL with optimization-based methods) have shown promising results.  However, there are still challenges in conventional methods, including edge mismatch and texture over-transferred between the RGB/depth images, difficulty to learn of natural priors effectively, and limited interpretability for the internal mechanism of DL architectures.

To this end, we propose a {\em Discrete Cosine Transform Network}~(DCTNet) for the GDSR task, inspired by coupled dictionary learning and physics-based modeling. It consists of four components: semi-coupled feature extraction~(SCFE), guided edge spatial attention~(GESA), discrete cosine transform~(DCT) module, and a depth reconstruction~(DR) module. The workflow is illustrated in Fig.~\ref{fig:Workflow}. Our contributions can be summarized as follows:

First, we propose the {\em semi-coupled} residual blocks to leverage the correlation between the intensity edge in RGB images and the depth discontinuities in depth images, but still preserve the unique properties like detailed texture and segment smoothness in two modalities. In each convolutional layer of this block, half of the kernels are responsible for extracting {\em shared} information in depth/RGB images, which is applied to both modalities. The rest half of the convolution kernels are designed to extract unique information in the depth and RGB images, respectively. Parameters in the {\em private} kernel are not shared. Thus the feature extractor with semi-coupled blocks can effectively extract informative features for GDSR from input image pairs.

Second, we propose a novel DCT module to improve the explainability of working mechanisms in the empirically-designed DL architectures.
This component utilizes DCT to solve a well-designed optimization model for GDSR and inserts it in the DL model as a module to acquire HR depth map features guided by RGB features in the multi-channel feature domain. Therefore, besides learning the LR-to-HR mapping, our DCTNet focuses more on feature extraction and edge weight highlighting. Although recent works have used DCT for recognition~\cite{xu2020learning} and image SR~\cite{magid2021dynamic}, we are the first to use it in restoring degraded depth maps to the best of our knowledge. We further make the tuning parameters in the DCT module learnable to improve model flexibility. 

Third, to overcome the issue that texture details in RGB images are over-transferred, we employ the enhanced spatial attention~(ESA) block from RFANet~\cite{DBLP:conf/cvpr/Liu0T0W20} in our GESA module to highlight the edges in RGB features useful for GDSR.
In this way, part of the intensity edges is activated and associated with the depth discontinuities, achieving the adaptive transfer from the texture structure in guided images.

We conduct comprehensive evaluations on four popular RGBD datasets, including NYU v2~\cite{DBLP:conf/eccv/SilbermanHKF12}, Middlebury~\cite{DBLP:conf/cvpr/HirschmullerS07,DBLP:conf/cvpr/ScharsteinP07}, Lu~\cite{DBLP:conf/cvpr/LuRL14} and RGBDD~\cite{DBLP:conf/cvpr/HeZLBCZLL021}. The quantitative and qualitative results show that our DCTNet can achieve state-of-the-art performance in GDSR with a relatively small number of parameters.  
	
\section{Related Work}\label{sec:2}

Super-resolution is a basic computer vision topic with many sub-fields and numerous approaches. Here we only discuss methods for GDSR.

\subsection{Conventional GDSR methods}

\bfsection{Filter-based methods} The filter-based (local) methods aim to use the RGB image to guide the joint filter to perceive the edge in depth map. Starting from joint bilateral upsampling~\cite{DBLP:journals/tog/KopfCLU07} and its variants~\cite{DBLP:conf/cvpr/YangYDN07,DBLP:journals/tcyb/CamplaniMS13}, RGB images guide the acquisition of bilateral weights. Liu~\etal~\cite{DBLP:conf/cvpr/0001TT13} replace the Euclidean distance with the geodesic distances to maintain the discontinuities of the depth image. Weighted mode filter~\cite{DBLP:journals/tip/MinLD12}, guided filtering~\cite{DBLP:journals/pami/He0T13} and its variants~\cite{DBLP:conf/cvpr/LuSMLD12,DBLP:conf/cvpr/TanSP14} are also widely used in the upsampling process. Lu~\etal~\cite{DBLP:conf/cvpr/LuF15} use the smoothing method to process the image parts obtained by the depth map guided RGB image segmentation to solve the texture transferring issue.

\bfsection{Optimization-based methods} The optimization-based (global) methods model the interdependency between color images and depth maps by Markov Random Field~\cite{DBLP:conf/nips/DiebelT05}, nonlocal means filtering~\cite{DBLP:conf/iccv/ParkKTBK11}, pixel-wise adaptive auto-regressive model~\cite{DBLP:journals/tip/YangYLHW14}, total generalized variation~\cite{DBLP:conf/iccv/FerstlRRRB13} and multi-pass optimization framework~\cite{DBLP:conf/eccv/LiMDL16}, respectively.

\bfsection{Learning-based methods} Earlier methods like bimodal co-sparse analysis~\cite{DBLP:conf/iccv/KiechleHK13} and joint dictionaries learning~\cite{DBLP:journals/tip/TosicD14} capture the interdependency of the RGB and depth images. A multi-scale dictionary learning strategy with RGB-D structural similarity measure and a robust coupled dictionary learning algorithm with local coordinate constraints are employed by Kwon~\etal~\cite{DBLP:conf/cvpr/KwonTL15} and Xie~\etal~\cite{DBLP:journals/tmm/XieFYS15} to solve over-smoothing and over-fitting problems in information transfer, respectively.
Gu~\etal~\cite{DBLP:conf/cvpr/GuZGCCZ17} establish a task-driven learning method to learn the dynamic guidance by a weighted analysis representation model. Xie~\etal~\cite{DBLP:journals/tip/XieFS16} learn an HR edge map inference method from external HR/LR image pair.

\subsection{Deep Learning GDSR Methods}
GDSR performance is further promoted with the powerful feature extraction capability of neural networks. Riegler~\etal~\cite{DBLP:conf/bmvc/RieglerFRB16} adopt the first-order primal-dual algorithm and unroll the optimization processing to a network structure, establishing the relationship between the DL-based methods and the optimization-based methods. Li~\etal~\cite{DBLP:conf/eccv/LiHA016,DBLP:journals/pami/LiHAY19} use a two-stream end-to-end network with skip connection to learn the mapping of LR to HR depth maps. Hui~\etal~\cite{DBLP:conf/eccv/HuiLT16} propose multi-scale guidance for edge transfer. Similarly, Guo \etal~\cite{DBLP:journals/tip/GuoLGCFH19} use the residual U-Net structure to learn the residual information between bicubic interpolation upsampling and ground truth under multi-scale guidance. CoIAST~\cite{DBLP:journals/tip/DengD20}, which is based on the iterative shrinkage thresholding algorithm~(ISTA)~\cite{DBLP:conf/icml/GregorL10}, regard the estimation of HR depth map as a linear combination of two LISTA branches.
CU-Net~\cite{DBLP:journals/pami/0002D21} uses two modules to separate common/unique features by multi-modal convolutional sparse coding and elaborate the model interpretability.
More recently, DKN~\cite{DBLP:journals/ijcv/KimPH21} and FDSR~\cite{DBLP:conf/cvpr/HeZLBCZLL021} achieve adaptive filtering neighbors/weight calculation and high-frequency guided feature decomposition through spatially-variant kernels learning and octave convolution, respectively. They outperform previous state-of-the-art~(SOTA) methods in synthetic and real scene datasets.

\subsection{Comparison with existing approaches}
Our proposed DCTNet is closely related to the optimization-based and DL-based coupled dictionary learning methods. (1)~The DCT module in our model obtains the depth map features of HR by solving an optimization problem, and we are the first to use DCT to solve the problem to our knowledge.
In addition, the DCT module is integrated into the DL framework to complete the multi-channel feature acquisition. The learnable parameters further enhance the flexibility of the optimization function in this module.
(2)~For the RGB texture over-transferred challenge, compared to local/global methods, we use the ESA module~\cite{DBLP:conf/cvpr/Liu0T0W20} to adaptively learn the edges attention weights in a data-driven manner.
(3)~Our feature extraction encoder is inspired by coupled dictionary learning, but we do not need to learn the dictionary explicitly. Instead, the private/shared feature extraction is accomplished by limiting whether the parameters are shared between the convolution kernels.
	
	\begin{figure*}[t]
		\centering
		\includegraphics[width=1.0\linewidth]{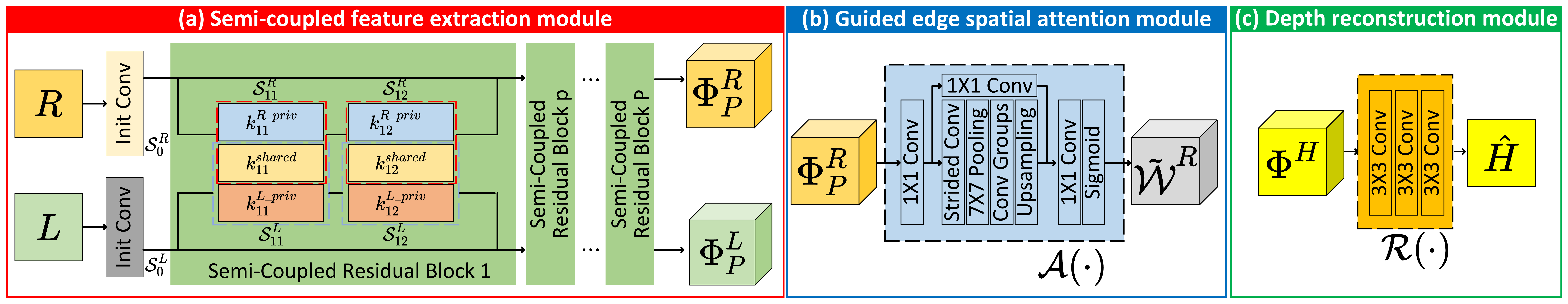}
		\vspace{-2em}
		\caption{Detailed illustration of DCTNet workflow. Sub-figures (a)-(c) are the specific structures of SCFE, GESA, and DR modules in Fig.~\ref{fig:Workflow}, which aims to extract cross-modality features, highlight RGB edge information, and reconstruct the HR depth map, respectively.}
		\label{fig:detail}
	\end{figure*}

	\section{Method}\label{sec:3}
    In this section, we will elaborate on the details of our proposed DCTNet. We first show how to use discrete cosine transform (DCT) to solve an optimization problem for the GDSR task in the image domain. We then describe the architectural units and training objectives of DCTNet.
	
	\subsection{Problem formulation}
    We first define some important symbols for clarity. In the GDSR task, a model is expected to take the HR RGB image $R\!\in\!\mathbb{R}^{M\times N\times 3}$ and the LR depth image $\tilde L\!\in\!\mathbb{R}^{m\times n}$ as inputs, where $\{M,N\}$ and $\{m,n\}$ are the height and width of input RGB and depth images, respectively. We aim to obtain the HR depth image $H\!\in\!\mathbb{R}^{M\times N}$ under the guidance of $R$. We also perform some preprocessing to get $\tilde R$ and $L$, where $\tilde R\!\in\!\mathbb{R}^{M\times N}$ denotes the Y channel in YCrCb color space of $R$ and $L\!\in\!\mathbb{R}^{M\times N}$ is the upsampled image of $\tilde L$. If $R$ and $\tilde L$ in the same scene are given, $H$ can be obtained by minimizing the following energy function:
	\begin{equation}\label{eq:optimization}
		\mathcal{F}=\frac{1}{2}\|H-L\|_{2}^{2}+\frac{\lambda}{2}\|\mathcal{L} (H)-\mathcal{L}(\tilde R) \circ \mathcal{W}(\tilde R)\|_{2}^{2},
	\end{equation}
	where $\mathcal{L}(\cdot)$ is the Laplacian filter, $\mathcal{W}(\cdot)$ can be regarded as a given threshold function to select the edges useful for GDSR. $\circ$ denotes element-wise multiplication, and $\lambda$ is a parameter controlling the contribution of the second term. 
	The optimal solution can be achieved when $\frac{\partial \mathcal{F}}{\partial H}=0$, and we have
	\begin{equation}\label{eq:DCT}
		H+\lambda \mathcal{L}^{2}(H)=\lambda \mathcal{L}\left(\mathcal{L}(\tilde R)\circ \mathcal{W}(\tilde R)\right)+L.
	\end{equation}	
	Eq.~(\ref{eq:DCT}) can be treated as a 2D Poisson's equation (PE). Here we assume a ``reflection padding'' extension at the boundary of the image when performing convolution operations, which makes zero gradients on the image boundary. Thus, PE Eq.~(\ref{eq:DCT}) is with the Neumann boundary condition~(NBC). Technically, PE with the NBC can be solved via DCT~\cite{strang1999discrete}. Then we set 
	$\lambda \mathcal{L}(\mathcal{L}(\tilde R)\circ \mathcal{W}(\tilde R))+L\triangleq E$, and implement DCT operation on both sides of the equation: 
	\begin{equation}
		\mathcal{F}_{c}(H)+\lambda K^{2} \circ \mathcal{F}_{c}(H)=\mathcal{F}_{c}(E),
	\end{equation}
	where $\mathcal{F}_{c}(\cdot)$ is the DCT operation, $K_{ij}\!=\!\cos \left(\frac{i-1}{M} \pi\right)+\cos \left(\frac{j-1}{N} \pi\right)$ and $1 \leq i \leq M,1 \leq j \leq N$. Finally, the HR depth images can be calculated by: 
	\begin{equation}\label{eq:DCT2}
		H=\mathcal{F}_{c}^{-1}\left\{ {\mathcal{F}_{c}(E)}\oslash\left(I+\lambda K^{2}\right) \right\},
	\end{equation}
	where $\mathcal{F}_{c}^{-1}(\cdot)$ is the inverse DCT operation, $\oslash$ denotes the element-wise division, and $I$ is the identity matrix. Due to space limitations, we refer readers to the supplementary material for the detailed derivation of the equation.
	
	The above method has the following problems:
	(a)~Although $H$ can be solved by optimization, it requires additional edge perception methods to determine $\mathcal{W}(\cdot)$. (b)~The $\lambda$ is manually given in Eq.~(\ref{eq:DCT}), which restricts the model flexibility. (c)~Optimizing a single channel in the image domain is difficult to effectively model the cross-modal internal feature correlation. 
	Combining the challenges discussed in Sec.~\ref{sec:1}, \eg, RGB texture over-transferred and the difficulty of natural prior learning, we propose a novel DCTNet in the following part to alleviate the above issues.
	
	\subsection{DCTNet}
	Our proposed DCTNet consists of four components including \textit{semi-coupled feature extraction}~(SCFE), \textit{guided edge spatial attention}~(GESA), \textit{discrete cosine transform}~(DCT) and \textit{depth reconstruction}~(DR) modules.
	The detailed illustrations are shown in Fig.~\ref{fig:Workflow} and \ref{fig:detail}.
	
	We give an overview of the model. First, given a pair of $L$ and $R$, the semi-coupled residual blocks extract shared and private features from the source images. The GESA module then processes the RGB feature to obtain the attention edge weights useful for SR. Subsequently, multi-channel RGB and depth features and the attention edge weights are input into the DCT module to acquire HR depth features. Finally, the depth reconstruction module outputs the SR depth map. The details of the modules are explained next.
	
	\subsubsection{Semi-coupled feature extraction}
    The RGB and depth maps in the same scene can have redundant information (\eg, shape and edges) and complementary information (\eg, RGB texture details and depth discontinuities). At the same time, based on the basic assumptions of the GDSR that some of the features in the cross-modal image should be interdependent, while other are modality-specific. Therefore, our SCFE module is designed to achieve cross-modal extraction of shared and private features.
	
    As shown in Fig.~\ref{fig:detail}(a), we can build the SCFE module as an encoder for feature extraction. The internal convolutions are two initial convolutions and $P$ semi-coupled residual blocks. Here we denote the initial convolution layers corresponding to $\{L,R\}$ as $\{\mathcal{S}^L_{0},\mathcal{S}^R_{0}\}$, and the $q$th convolution layer in the $p$th semi-coupled residual block corresponding to $\{L,R\}$ is denoted as $\{\mathcal{S}^L_{pq},\mathcal{S}^R_{pq}\}$, where $p=1,2,\cdots,P$ and $q=1,2$. The output features of $\{\mathcal{S}^L_{pq},\mathcal{S}^R_{pq}\}$ are denoted by $\{\Phi^L_{pq},\Phi^R_{pq}\}\in \mathbb{R}^{M\times N\times C}$, where $C$ is the number of kernels in $\{\mathcal{S}^L_{pq},\mathcal{S}^R_{pq}\}$. $P$ and $C$ are determined in Sec~\ref{sec:4_2}. Note that when $q=2$, $\{\Phi^L_{pq},\Phi^R_{pq}\}$ can be simplified as $\{\Phi^L_{p},\Phi^R_{p}\}$. The initialization layer generates $\Phi^R_{0}=\mathcal{S}^R_{0}(R)$, $\Phi^L_{0}=\mathcal{S}^L_{0}(L)$. 
    Then taking the first convolution kernel in the $p$th semi-coupled residual block as an example, the semi-coupled convolution operation can be expressed as
	\begin{align}
		\mathcal{S}^R_{p1}(\Phi ^R_{p-1})&=\Phi ^R_{p-1}*\mathcal{C}(k_{p1}^{shared},k_{p1}^{R\_priv}),\\
		\mathcal{S}^L_{p1}(\Phi ^L_{p-1})&=\Phi ^L_{p-1}*\mathcal{C}(k_{p1}^{shared},k_{p1}^{L\_priv}),
	\end{align}
	where $*$ denotes convolution, $\{k_{p1}^{shared},k_{p1}^{R\_priv},k_{p1}^{L\_priv}\}$ denote the shared convolution kernels and the private ones corresponding to $R$ and $L$, respectively. $\mathcal{C}(\cdot,\cdot)$ denotes the concatenation over the channel dimension. Then, the output feature $\Phi^R_{p}$ of $R$ in the $p$th residual block becomes
	\begin{equation}\label{eq:update_r}
		\Phi ^R_p=\mathtt{ReLU}\left\lbrace \mathcal{S}^R_{p2}(\mathtt{ReLU}(\mathcal{S}^R_{p1}(\Phi ^R_{p-1})))+\Phi ^R_{p-1}\right\rbrace,
	\end{equation}
	and that of $\Phi^L_{p}$ is similar to Eq.~(\ref{eq:update_r}), only the superscript needs to be replaced from $R$ to $L$. Finally, the outputs of SCFE module are $\Phi^L_{P}$ and $\Phi^R_{P}$, which contain both the shared and the private features in the cross-modal image pair.
	
    Compared with fully-shared or independent settings, the semi-coupled convolution kernels in the SCFE module can learn the shared/private parts of their respective input features, which extract features more effectively. The effectiveness of the SCFE module is demonstrated in Sec.~\ref{sec:Ablation}.
	
	\subsubsection{Guided edge spatial attention}
	To prevent the problem that irrelevant textures are transferred to the SR depth map $H$ when the guide RGB image contains rich textures, we adopt the ESA block from RFANet~\cite{DBLP:conf/cvpr/Liu0T0W20} that achieves excellent results in single image SR into our GESA module, as shown in Fig.~\ref{fig:detail}(b). 
	The ESA block can highlight the attention weight in a lightweight and efficient manner, facilitating the learning of discriminative features. This motivation meets our requirements for the GESA module.
	We use $ \mathcal{A} (\cdot)$ to represent the operation in this module, and the guided edge attention weight can be obtained by 
	\begin{equation}\label{eq:weight_attention}
		\tilde{\mathcal{W}}^R=\mathcal{A}(\Phi^R_{P})\in\mathbb{R}^{M\times N\times C}.
	\end{equation}
	
	This module replaces the operation of obtaining $\mathcal{W}(\tilde{R})$ by manually giving $\mathcal{W}(\cdot)$ in \cref{eq:optimization}. Thus, part of the edges in intensity features can be highlighted.
	Compared with conventional methods that manually design criteria to extract edge weights useful for upsampling, data-driven strategies can achieve adaptive extraction of attention weights.
	
	\subsubsection{Discrete cosine transform}
	In the above subsections, we have acquired the multi-channel features $\Phi^R$ and $\Phi^L$ corresponding to $R$, $L$\footnote{We denote $\{\Phi^L_{P},\Phi^R_{P}\}$ as $\{\Phi^L,\Phi^R\}$ for simplicity.}, and guided edge attention weight $\tilde{\mathcal{W}}^R$. In this subsection, we will use them to accomplish the depth feature upsampling. 
	In Eq.~(\ref{eq:DCT2}), we illustrate that given a pair of $L$, $R$ and a threshold functions $\mathcal{W}(\cdot)$, the HR depth image can be reconstructed through DCT operation. 
	Thus we consider the DCT algorithm as a module, which can be integrated into our DCTNet framework. Furthermore, it can be expanded to obtain the multi-channel HR depth map features by completing the DCT operation on each feature channel. Mathematically, the calculation of the DCT module, denoted as $\mathcal{DCT}(\cdot,\cdot,\cdot)$, is 
	\begin{equation}\label{eq:DCT4}
		\Phi^H=\mathcal{DCT}(\Phi^R,\Phi^L,\tilde{\mathcal{W}}^R),
	\end{equation}
	where $\Phi^H\in \mathbb{R}^{M\times N\times C}$ is the guided upsampling feature of depth map $L$. More specifically, $\mathcal{DCT}(\cdot,\cdot,\cdot)$ calculates
	\begin{align}\label{eq:DCT3}
		\Phi^E[c]&\triangleq\tilde{\lambda}_c \mathcal{L}\left(\mathcal{L}(\Phi^R[c])\circ \tilde{\mathcal{W}}^R[c]\right)+\Phi^L[c],\\
		\Phi^H[c]&=\mathcal{F}_{c}^{-1}\left\{ {\mathcal{F}_{c}(\Phi^E[c])}\oslash\left(I+\tilde{\lambda}_c K^{2}\right) 	\right\},
	\end{align}
	where $\Phi^H[c]\!\in\!\mathbb{R}^{M\times N}$ is the $c$th channel feature map of $\Phi^H$. We want to emphasize that, compared with the manually given $\lambda$ in Eq.~(\ref{eq:optimization}) and Eq.~(\ref{eq:DCT2}), the $\tilde{\lambda}\!\in\!\mathbb{R}^C$ in Eq.~(\ref{eq:DCT3}) is set to be learnable. The channel-wise parameters are updated with the training progress, improving model flexibility.
	
    To summarize, there are two main advantages of using the DCT module. First, besides $\tilde{\lambda}\!\in\!\mathbb{R}^C$, the acquisition of the feature map $\Phi^H$ is learning-free, which can reduce the network size with less learnable weights. Second, using the DCT operation to directly calculate the output features makes this component more interpretable than a neural network that usually works like a black box.
	
	\subsubsection{Depth reconstruction}
	Finally, the depth reconstruction module aims to predict the HR depth map from its feature map $\Phi^H$, which is the output of the DCT module. The detailed structure is shown in Fig.~\ref{fig:detail}(c). Specifically, the function $\mathcal{R}(\cdot)$ of this module can be expressed as $\hat{H} = \mathcal{R}(\Phi^H)$, where $\hat{H}\!\in\!\mathbb{R}^{M\times N}$ is the predicted HR depth map of DCTNet.
	
	\subsubsection{Training loss}
	Consistent with recent works~\cite{DBLP:conf/eccv/LiHA016,DBLP:journals/pami/LiHAY19,DBLP:journals/pami/0002D21}, we choose $\ell_2$-loss as the training objective. That is, $\mathcal{D}(\hat{H}_i,H_i)=\sum_{i=1}^{N} \|\hat{H}_i-H_i\|_{2}^2$, where $H_i$ is the ground truth HR depth map. 
	\begin{figure*}[t]
		\begin{minipage}[b]{0.63\linewidth}
			\centering
			\includegraphics[width=0.8\linewidth]{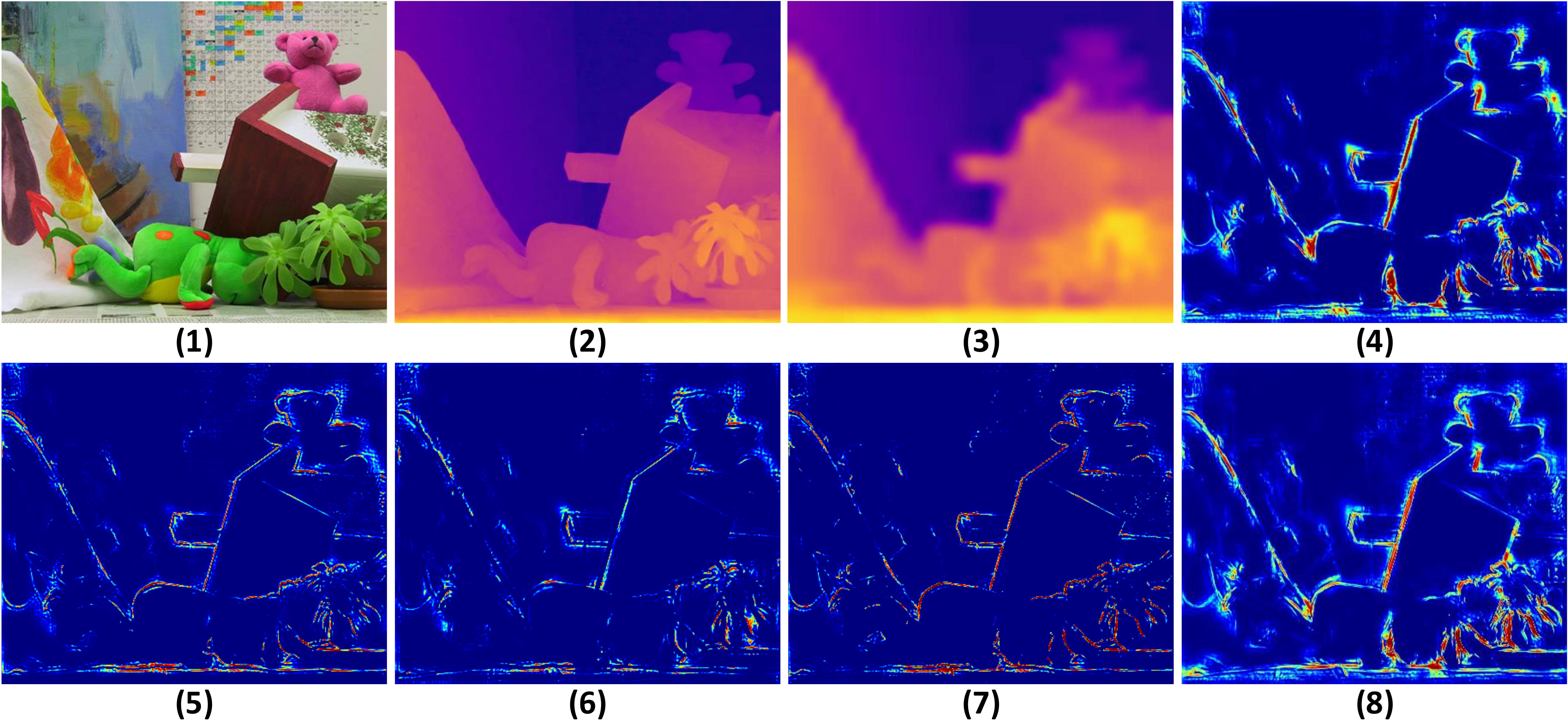}
			\par\vspace{0pt}
		\end{minipage}%
		\begin{minipage}[b]{0.37\linewidth}
			\centering
			\includegraphics[width=0.8\linewidth]{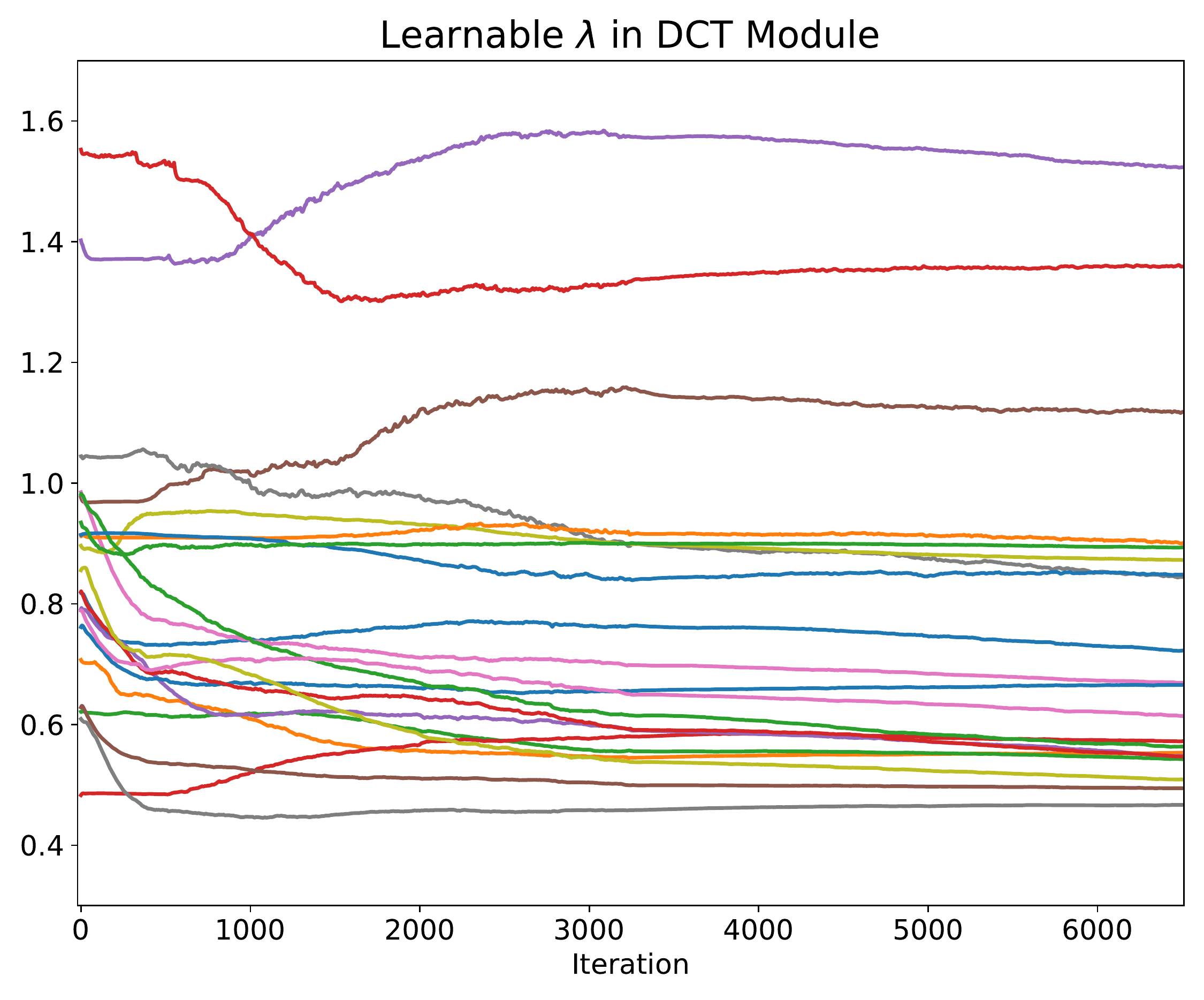}
			\par\vspace{0pt}
		\end{minipage}
		
		\vspace{-0.5em}
		\caption{
		Visual results of (left) highlighted edge attention weights and (right) the changing curves of learnable $\lambda$. Left: (1)-(3): Input $R$, ground truth $H$ and Input $L$, respectively. (4)-(8): Representative highlighted edge weights produced by the GESA module. Right: the values of the learnable parameters $\tilde\lambda$ against iterations during training. Different colored lines denote $\tilde\lambda_c$ corresponding to different channels.}
		\label{fig:learnable_lambda}
	\end{figure*}
	\begin{table}[t]
		\centering
		\resizebox{0.8\linewidth}{!}{
			\begin{tabular}{cccccc}
				\toprule
				  \multicolumn{6}{c}{\textbf{Impact of network depth $P$ ($C=64$)}}    \\
				  Setting   &   2   &   3   &       4        &       5        &       6        \\ \midrule
				$\times$4  & 2.378 & 1.989 &     1.544      & \textbf{1.521} &     1.527      \\
				$\times$8  & 4.644 & 3.963 & \textbf{3.152} &     3.174      &     3.166      \\
				$\times$16 & 8.245 & 6.904 & \textbf{5.764} &     5.787      &     5.776      \\ \midrule
				   \multicolumn{6}{c}{\textbf{Impact of network width $C$ ($P=4$)}}    \\
				  Setting   &   8   &  16   &       32       &       64       &      128       \\ \midrule
				$\times$4  & 2.798 & 2.300 &     1.992      &     1.544      & \textbf{1.529} \\
				$\times$8  & 5.694 & 4.476 &     3.808      & \textbf{3.152} &     3.171      \\
				$\times$16 & 9.531 & 7.695 &     6.976      &     5.764      & \textbf{5.734} \\ \bottomrule
			\end{tabular}}
		
		\vspace{-1em}
		\caption{The impacts of depth $P$ and width $C$ on the DCTNet using the validation set. \textbf{Bold} indicates the best RMSE result.}
		\label{tab:depth}
	\end{table}	
	\section{Experiment}\label{sec:4}
    In this section, we conduct comprehensive quantitative and qualitative experiments on several datasets to demonstrate the effectiveness of our proposed DCTNet.
	
	\subsection{Setup}
	\bfsection{Datasets}
    We use popular GDSR benchmarks following the protocol in recent works~\cite{DBLP:conf/eccv/LiHA016,DBLP:journals/pami/LiHAY19,DBLP:conf/cvpr/SuJSGLK19,DBLP:journals/ijcv/KimPH21}. Specifically, we select the first 1000 pairs of NYU v2 dataset~\cite{DBLP:conf/eccv/SilbermanHKF12} as the training set (900 pairs for training the network and 100 pairs for validation), and the last 449 pairs as a test set. We also utilize Middlebury~\cite{DBLP:conf/cvpr/HirschmullerS07,DBLP:conf/cvpr/ScharsteinP07} (30 pairs) and Lu~\cite{DBLP:conf/cvpr/LuRL14} (6 pairs) provided by Lu \etal~\cite{DBLP:conf/cvpr/LuRL14} as the test sets. In addition, 405 pairs of images in the RGBDD dataset~\cite{DBLP:conf/cvpr/HeZLBCZLL021} are incorporated for evaluation. We train our DCTNet on the NYU v2 dataset~\cite{DBLP:conf/eccv/SilbermanHKF12} and test on the four datasets mentioned above. 
	
    In our experiments, all the LR depth images are synthesized by applying bicubic down-sampling of the HR depth maps. Finally, to verify the generalization ability of our models in natural scenes, we test them on the \textit{real-world branch} of the RGBDD dataset\footnote{This branch dataset contains 2215/405 pairs of RGBD images as training/test set. The LR depth maps and target HR depth maps are all acquired in real scenes, with the sizes of 192$\times$144 and 512$\times$384, respectively.}~\cite{DBLP:conf/cvpr/HeZLBCZLL021}. Please refer to the supplementary material for more descriptions of all the datasets.
	
	\bfsection{Metric and implementation details}
    The training samples are resized to 256$\times$256 in the pre-processing stage. The network is trained for 1000 epochs with a mini-batch size of 64. We use the Adam~\cite{kingma2014adam} optimizer with a learning rate of $10^{-3}$. In the test phase, we follow common practice and use the root-mean-square error~(RMSE) to measure the depth SR performance against the ground-truth maps. A smaller RMSE implies a better quality of predicted depth images. The scripts are mainly implemented with Pytorch~\cite{NEURIPS2019_9015}. The training and testing are carried out on a PC with two NVIDIA GeForce RTX 3090 GPUs. We randomly initialize the learnable parameter $\tilde{\lambda}$ to $e^\theta$, where $\theta\sim \mathcal{N}(0.1,0.3)$. The number of semi-coupled residual blocks $P$ and the kernel numbers $C$ of semi-coupled filters in each convolution layer are set to 4 and 64, respectively. The choices of $P$ and $C$ are verified using the validation set in Sec.~\ref{sec:4_2}.

	\newcommand{\widthh}{0.175}
	\newcommand{\hhspace}{-0.1em}
	\newcommand{\lline}{0.999}
	\newcommand{\titlevspace}{-0.1em}
	\begin{figure*}[t]
		\centering
		\begin{subfigure}{\widthh \linewidth}
			\centering
			\includegraphics[width=\lline \linewidth]{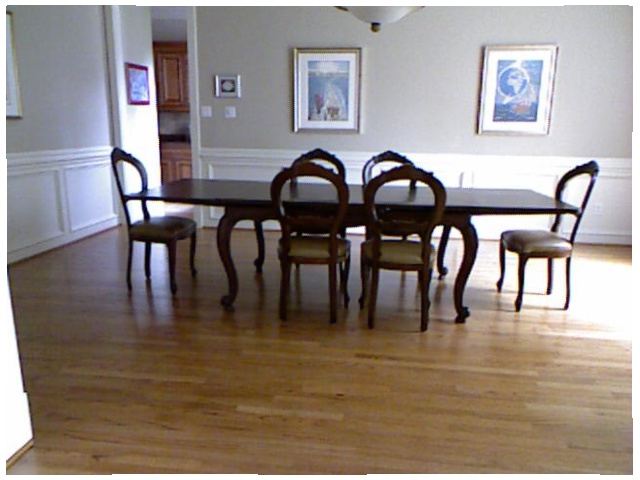}
			
			\vspace{\titlevspace}
			\caption*{(a) RGB}
		\end{subfigure}\hspace{\hhspace}
		\begin{subfigure}{\widthh \linewidth}
			\centering
			\includegraphics[width=\lline \linewidth]{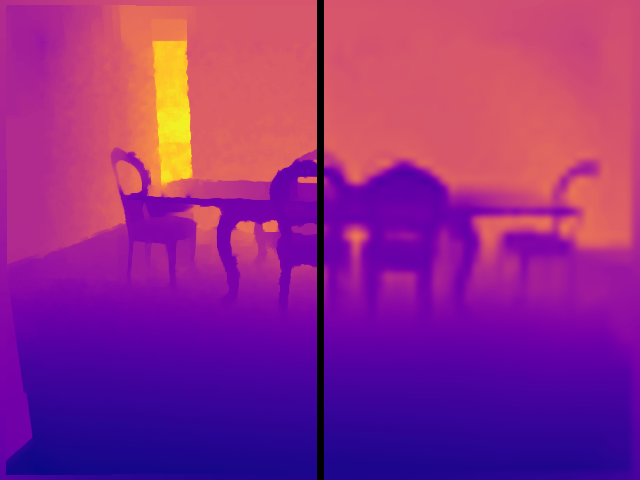}
			
			\vspace{\titlevspace}
			\caption*{(b) HR/LR Depth map}
		\end{subfigure}\hspace{\hhspace}
		\begin{subfigure}{\widthh \linewidth}
			\centering
			\includegraphics[width=\lline \linewidth]{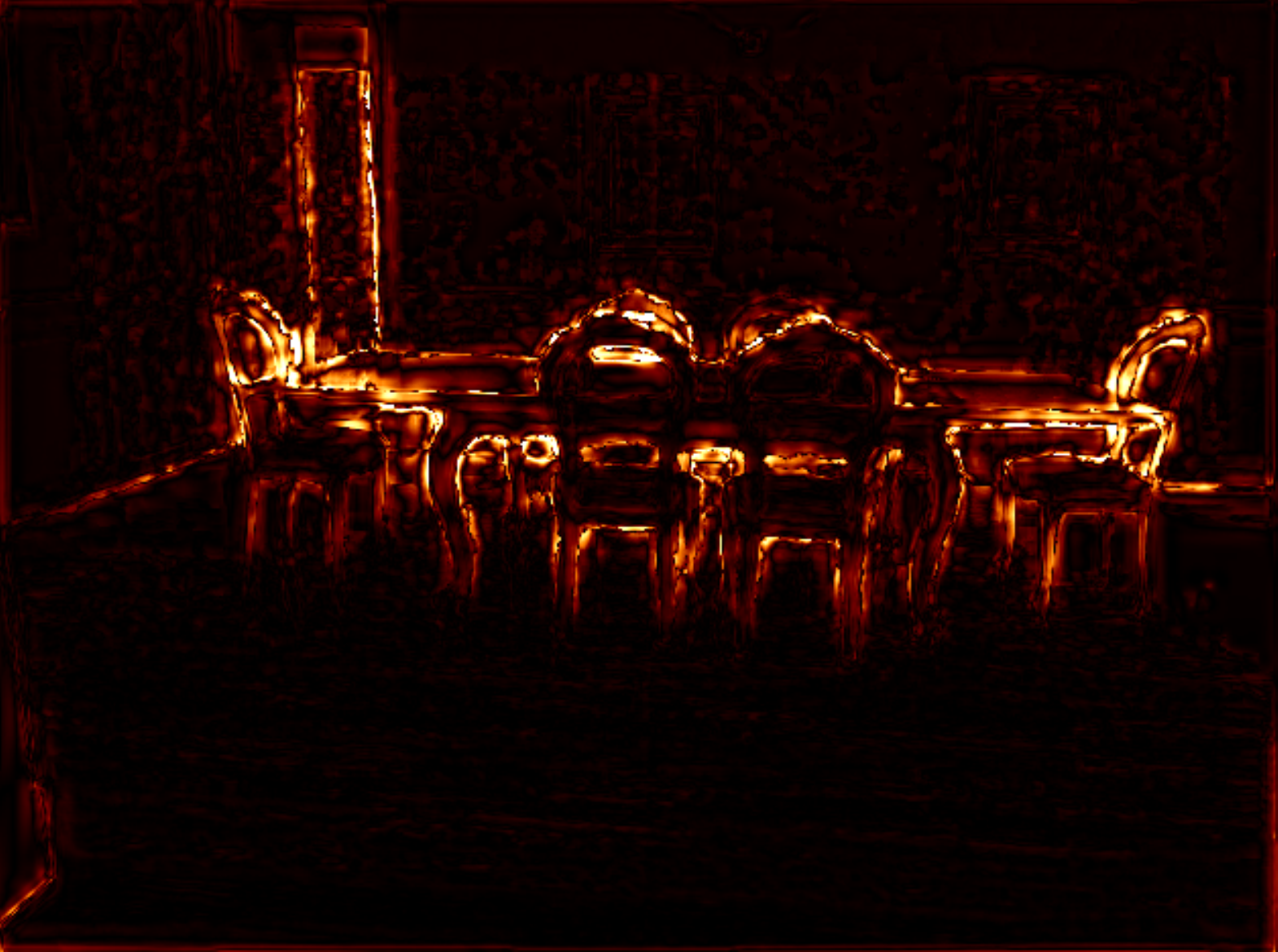}
			
			\vspace{\titlevspace}
			\caption*{(c) DJF~\cite{DBLP:conf/eccv/LiHA016}}
		\end{subfigure}\hspace{\hhspace}
		\begin{subfigure}{\widthh \linewidth}
			\centering
			\includegraphics[width=\lline \linewidth]{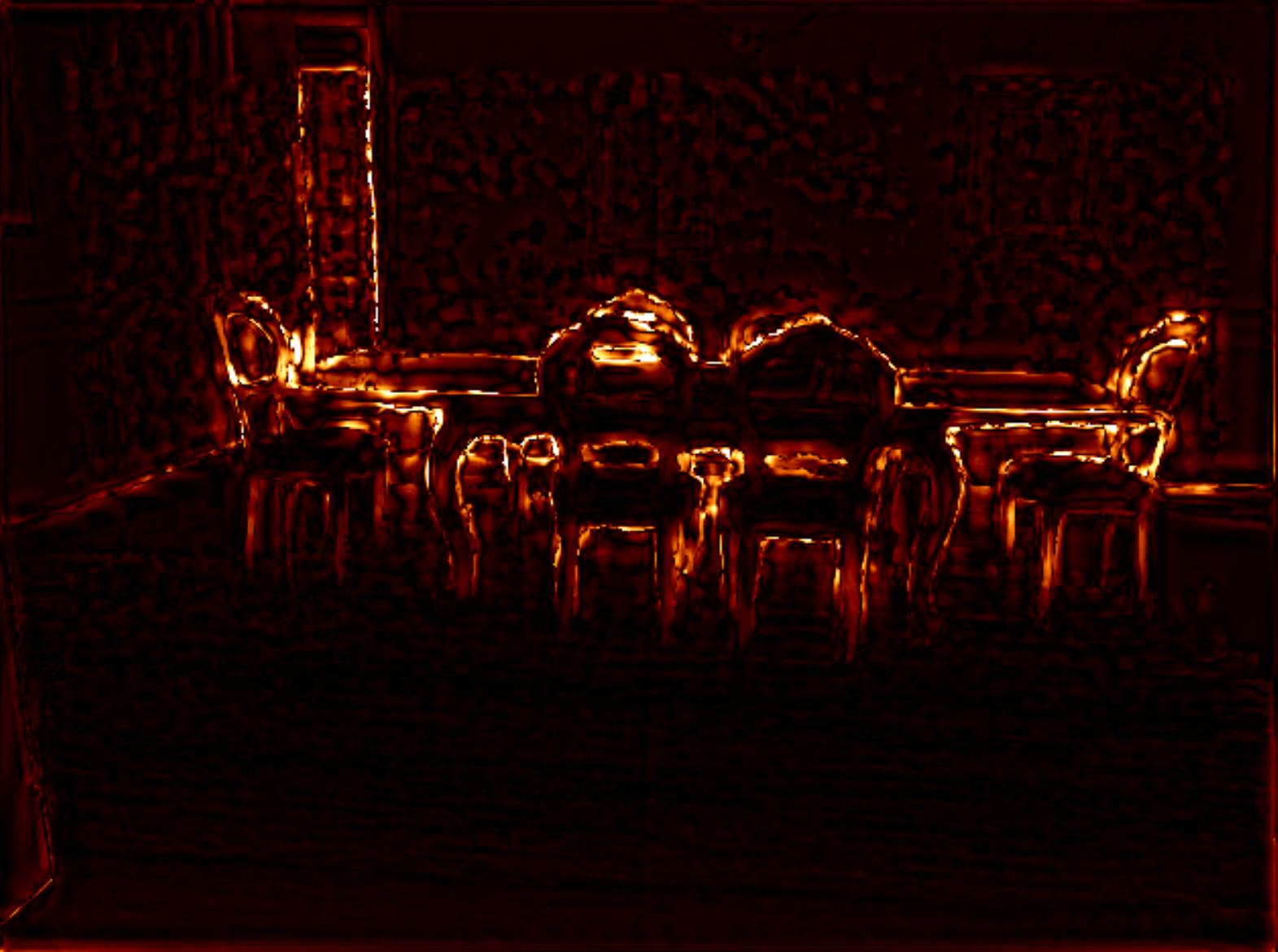}
			
			\vspace{\titlevspace}
			\caption*{(d) DJFR~\cite{DBLP:journals/pami/LiHAY19}}
		\end{subfigure}\hspace{\hhspace}
		\begin{subfigure}{\widthh \linewidth}
			\centering
			\includegraphics[width=\lline \linewidth]{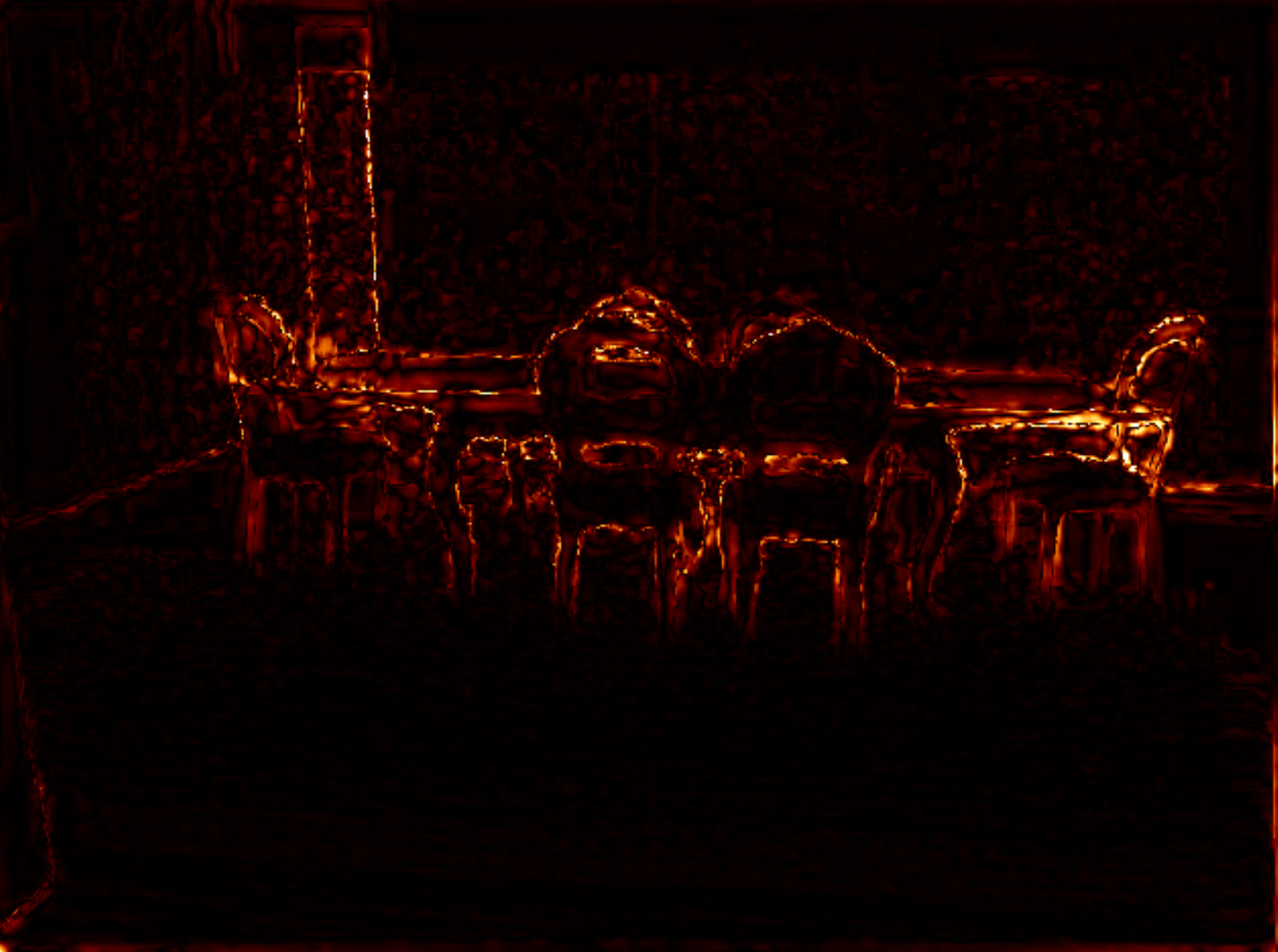}
			
			\vspace{\titlevspace}
			\caption*{(e) PAC~\cite{DBLP:conf/cvpr/SuJSGLK19}}
		\end{subfigure}	\hspace{\hhspace}
		
		\vspace{-0.1em}
		\begin{subfigure}{\widthh \linewidth}
			\centering
			\includegraphics[width=\lline \linewidth]{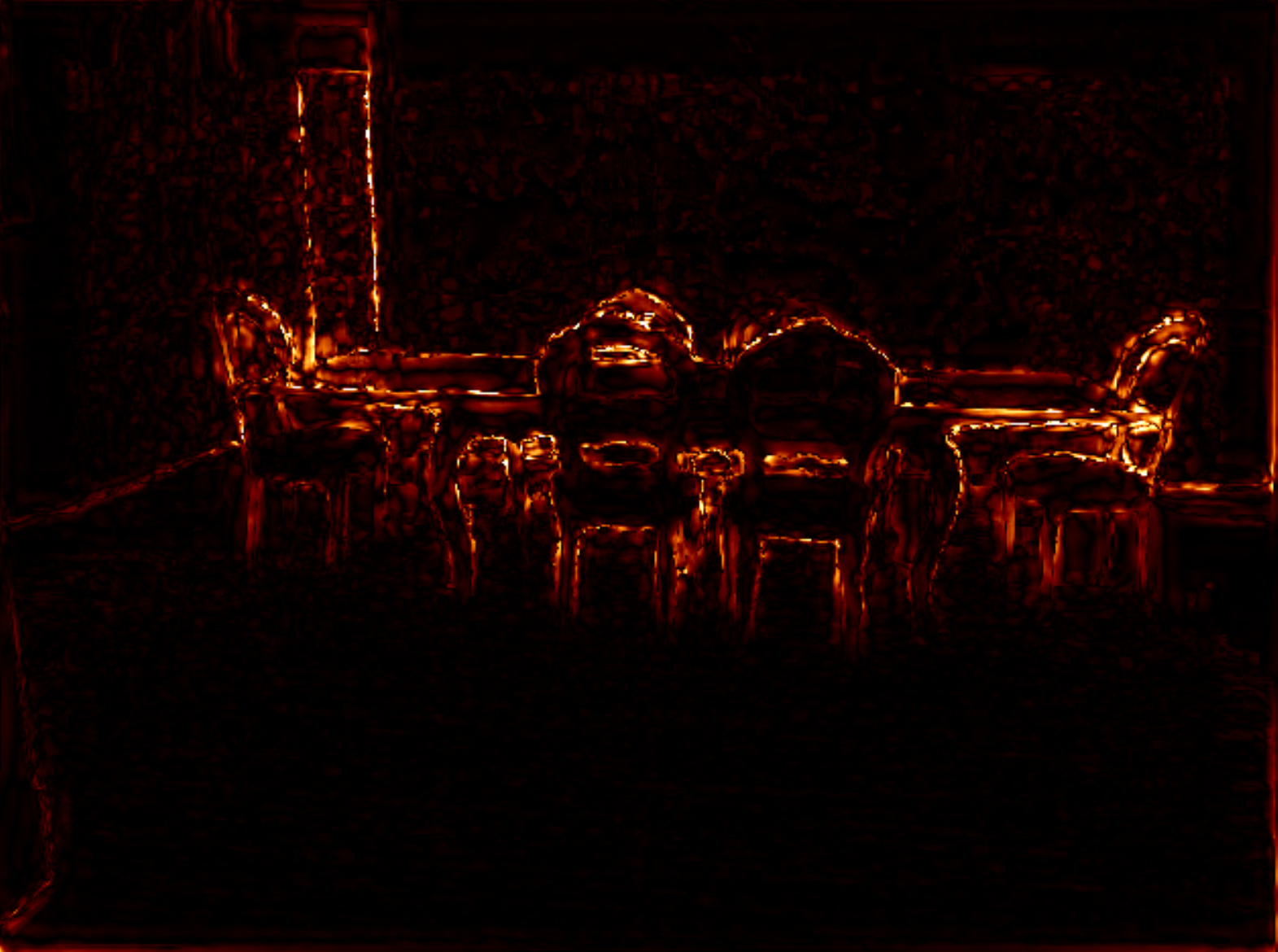}
			
			\vspace{\titlevspace}
			\caption*{(f) CUNet~\cite{DBLP:journals/pami/0002D21}}
		\end{subfigure}\hspace{\hhspace}
		\begin{subfigure}{\widthh \linewidth}
			\centering
			\includegraphics[width=\lline \linewidth]{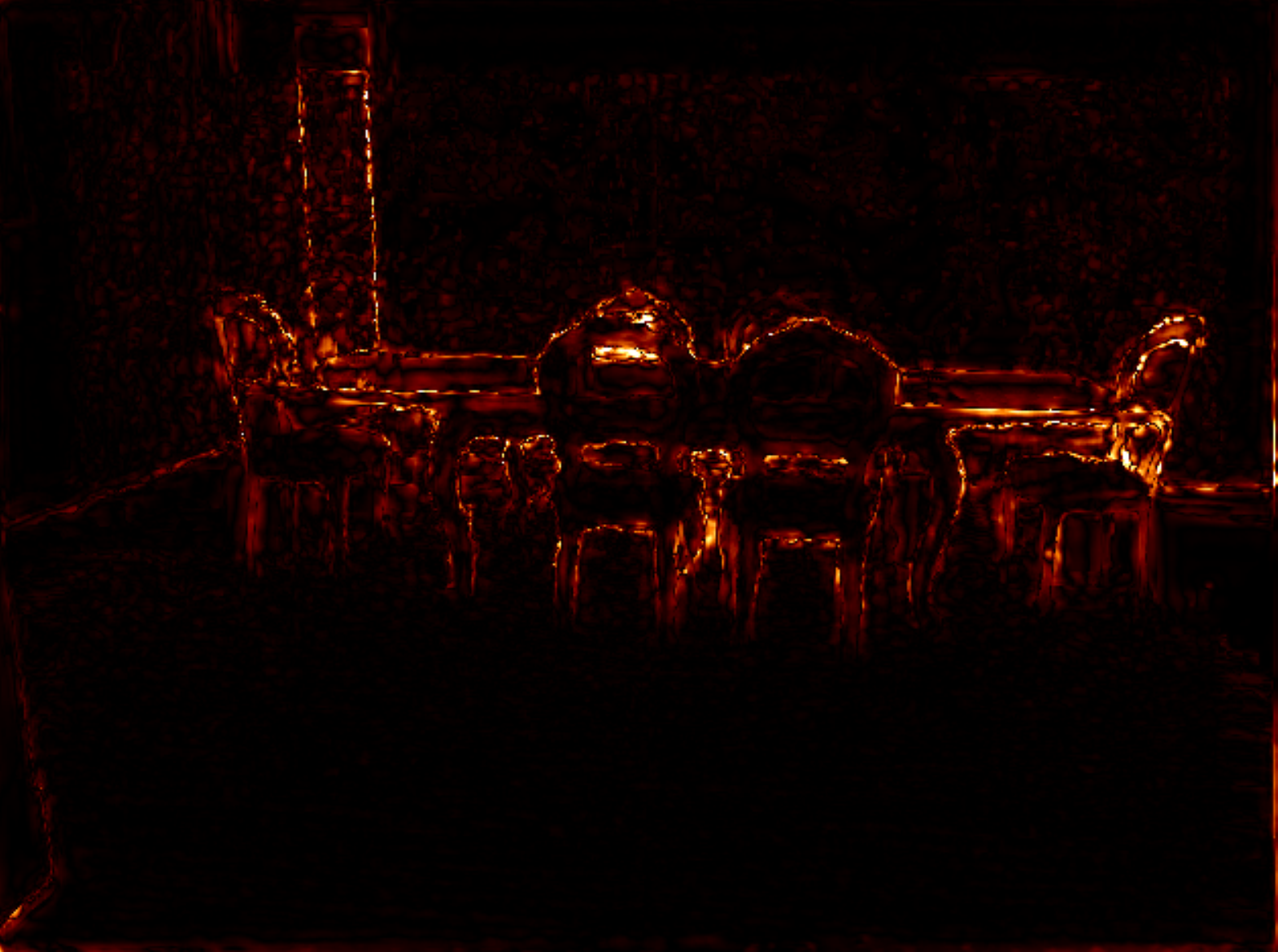}
			
			\vspace{\titlevspace}
			\caption*{(g) DKN~\cite{DBLP:journals/ijcv/KimPH21}}
		\end{subfigure}\hspace{\hhspace}
		\begin{subfigure}{\widthh \linewidth}
			\centering
			\includegraphics[width=\lline \linewidth]{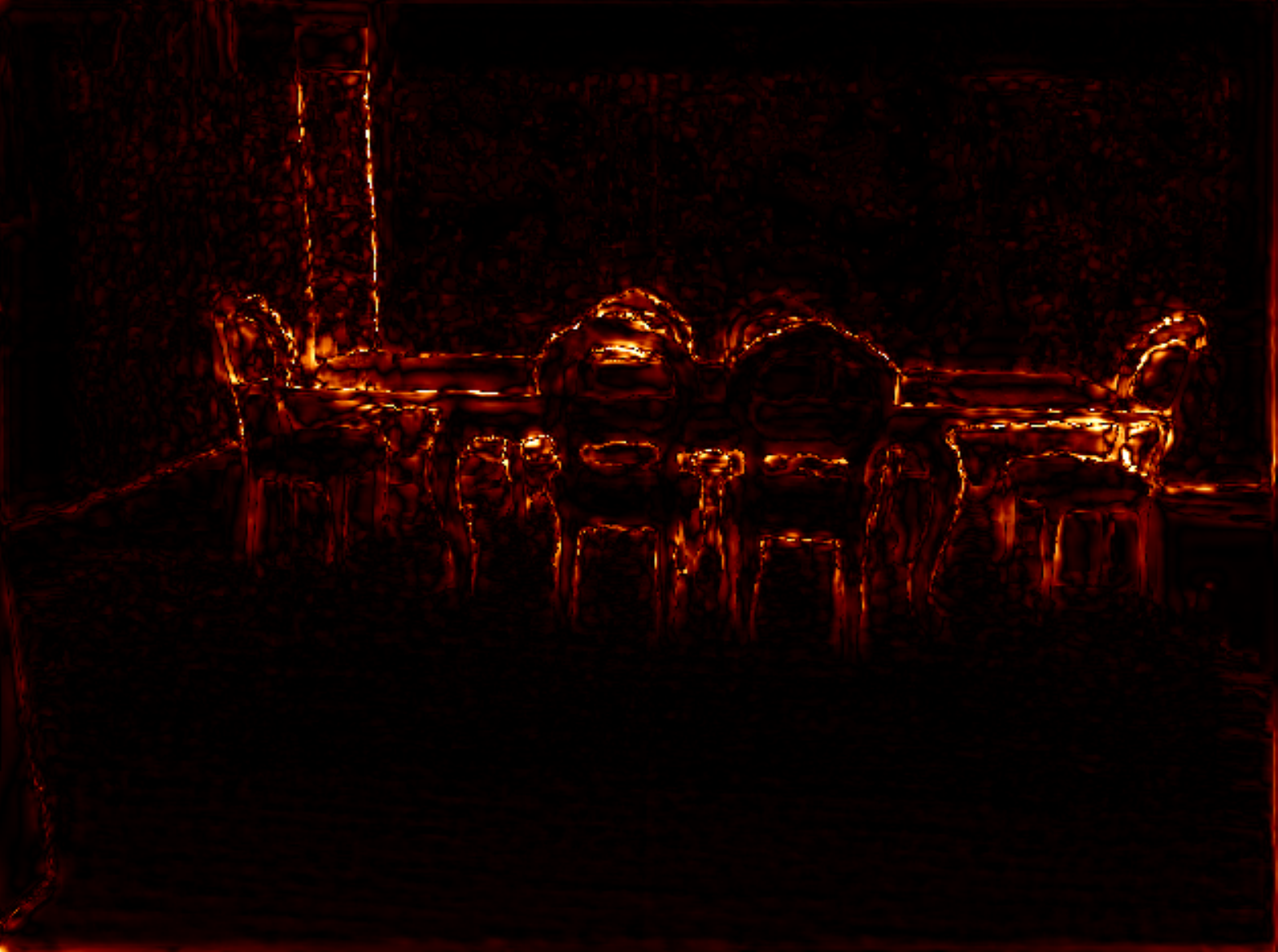}
			
			\vspace{\titlevspace}
			\caption*{(h) FDKN~\cite{DBLP:journals/ijcv/KimPH21}}
		\end{subfigure}\hspace{\hhspace}
		\begin{subfigure}{\widthh \linewidth}
			\centering
			\includegraphics[width=\lline \linewidth]{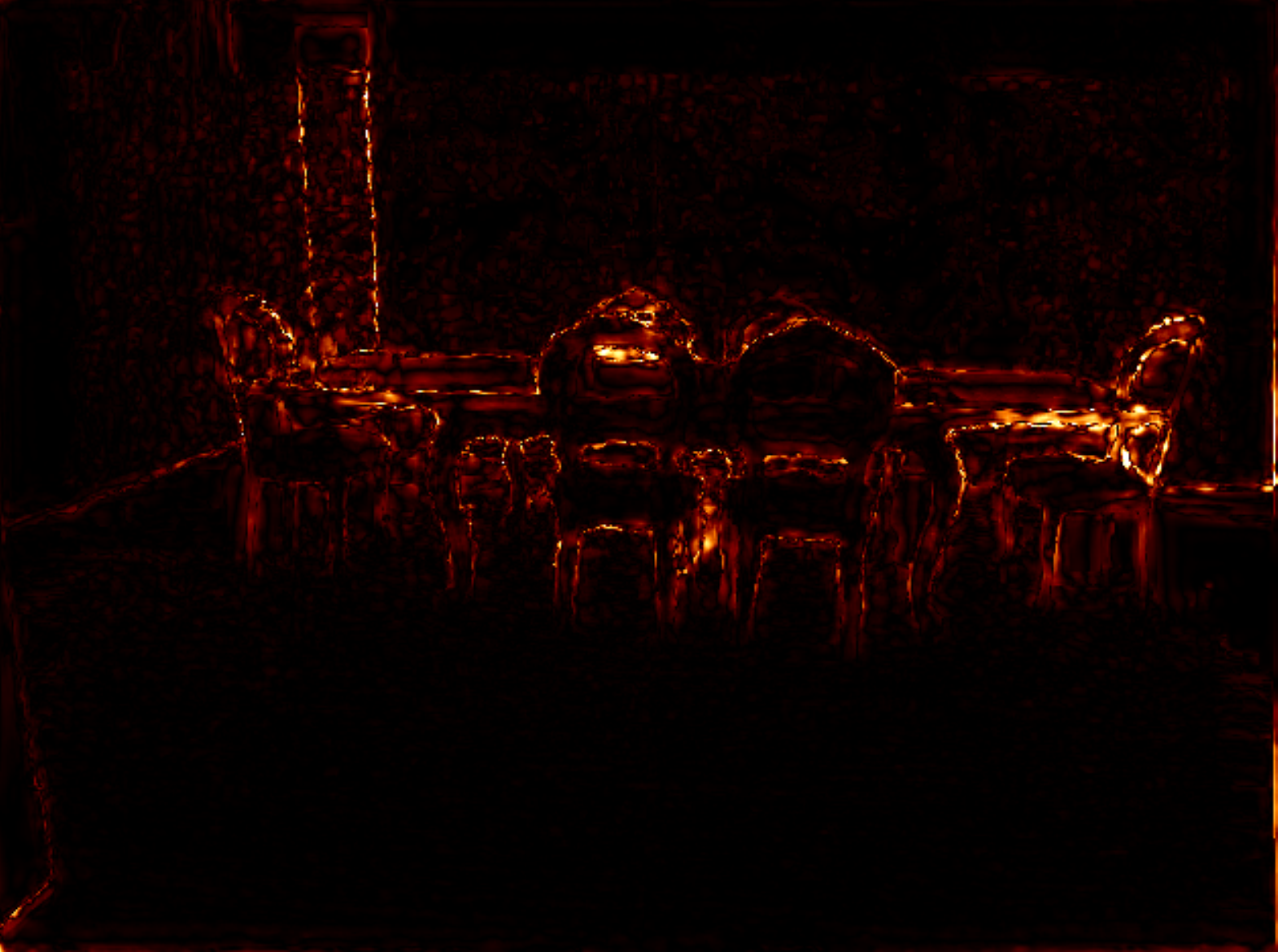}
			
			\vspace{\titlevspace}
			\caption*{(i) FDSR~\cite{DBLP:conf/cvpr/HeZLBCZLL021}}
		\end{subfigure}\hspace{\hhspace}
		\begin{subfigure}{\widthh \linewidth}
			\centering
			\includegraphics[width=\lline \linewidth]{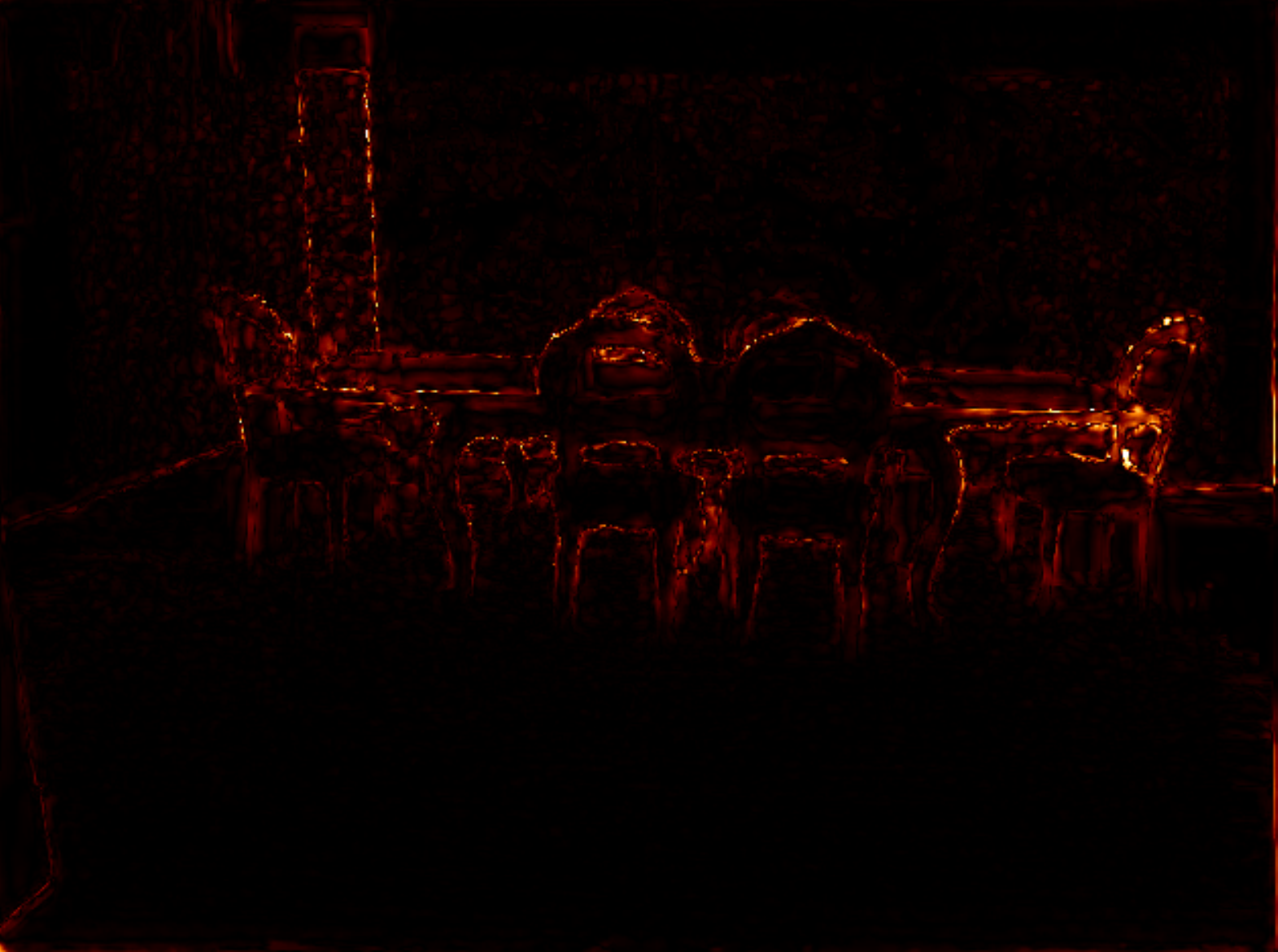}
			
			\vspace{\titlevspace}
			\caption*{(j) DCTNet ({\bf Ours})}
		\end{subfigure}	\hspace{\hhspace}
		
		\vspace{-1em}
		\caption{Visual comparison of error maps for ``Image\_1365'' in the NYU v2 dataset for 8$\times$ super-resolution.}
		\label{fig:Qualitative1}
	\end{figure*}

	\begin{figure*}[t]
		\centering
		\begin{subfigure}{\widthh \linewidth}
			\centering
			\includegraphics[width=\lline \linewidth]{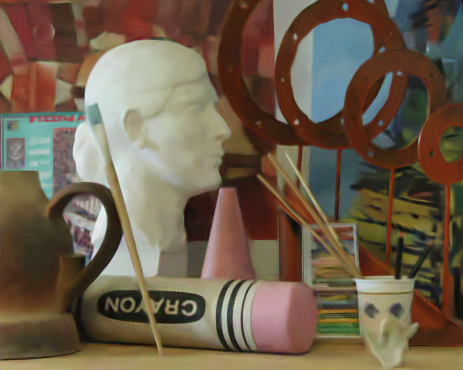}
			
			\vspace{\titlevspace}
			\caption*{(a) RGB}
		\end{subfigure}\hspace{\hhspace}
		\begin{subfigure}{\widthh \linewidth}
			\centering
			\includegraphics[width=\lline \linewidth]{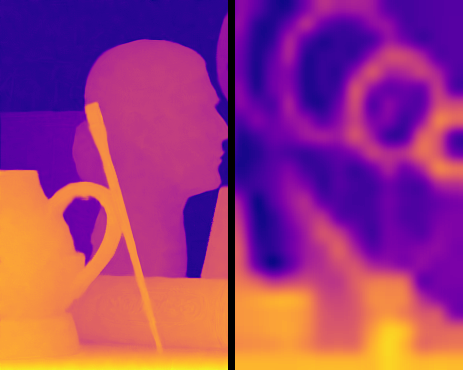}
			
			\vspace{\titlevspace}
			\caption*{(b) HR/LR Depth map}
		\end{subfigure}\hspace{\hhspace}
		\begin{subfigure}{\widthh \linewidth}
			\centering
			\includegraphics[width=\lline \linewidth]{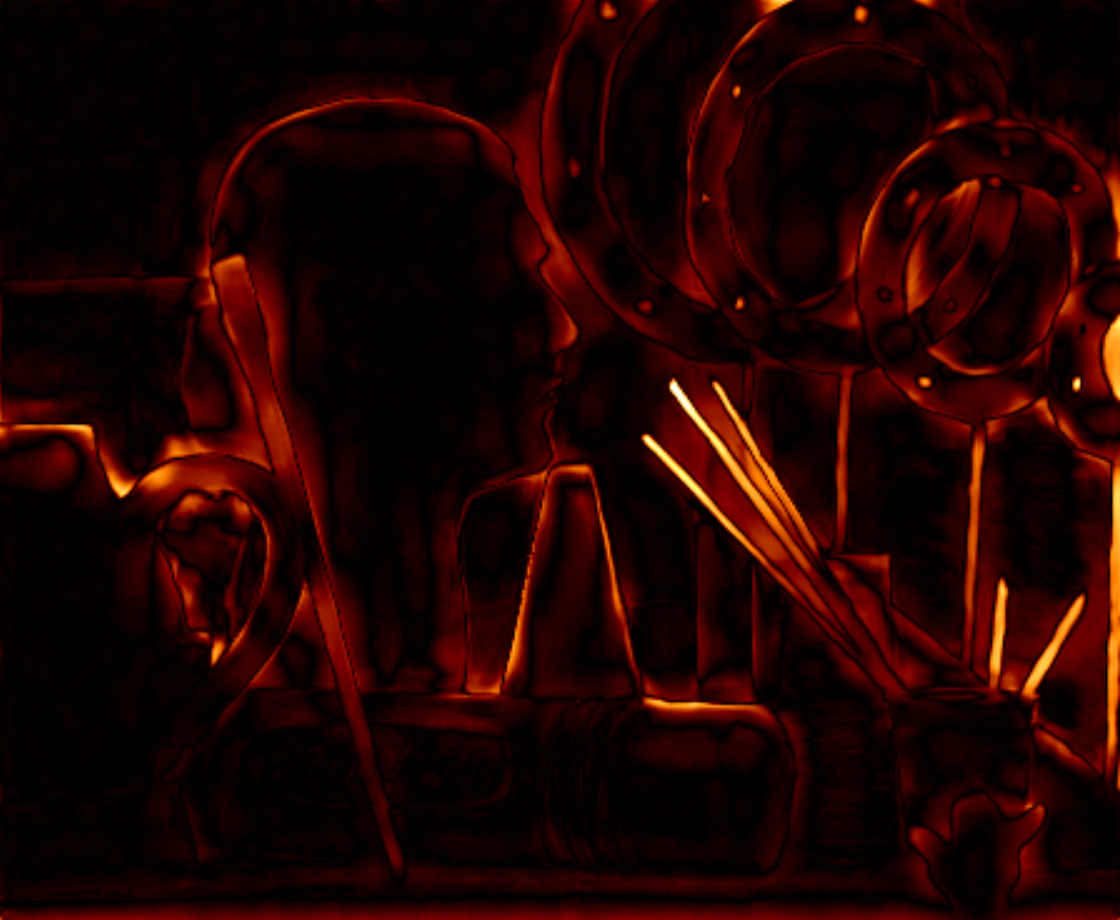}
			
			\vspace{\titlevspace}
			\caption*{(c) DJF~\cite{DBLP:conf/eccv/LiHA016}}
		\end{subfigure}\hspace{\hhspace}
		\begin{subfigure}{\widthh \linewidth}
			\centering
			\includegraphics[width=\lline \linewidth]{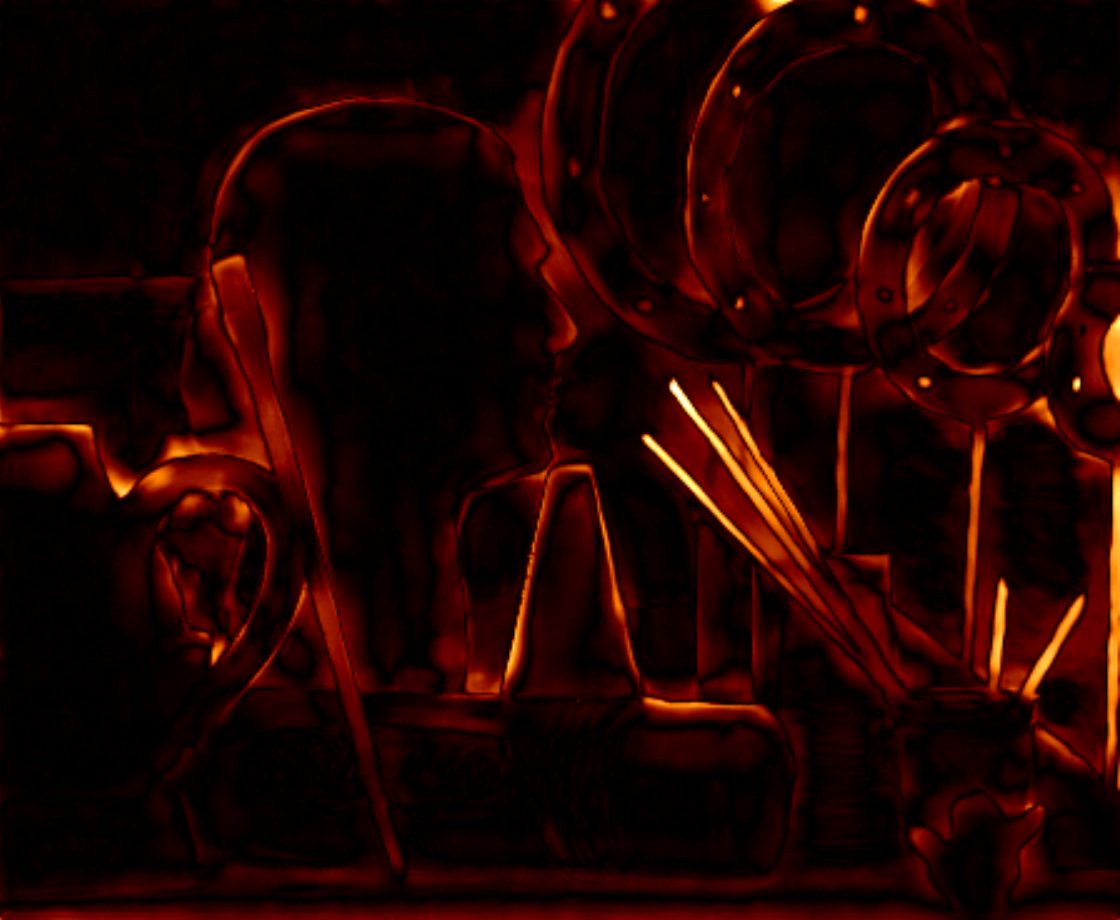}
			
			\vspace{\titlevspace}
			\caption*{(d) DJFR~\cite{DBLP:journals/pami/LiHAY19}}
		\end{subfigure}\hspace{\hhspace}
		\begin{subfigure}{\widthh \linewidth}
			\centering
			\includegraphics[width=\lline \linewidth]{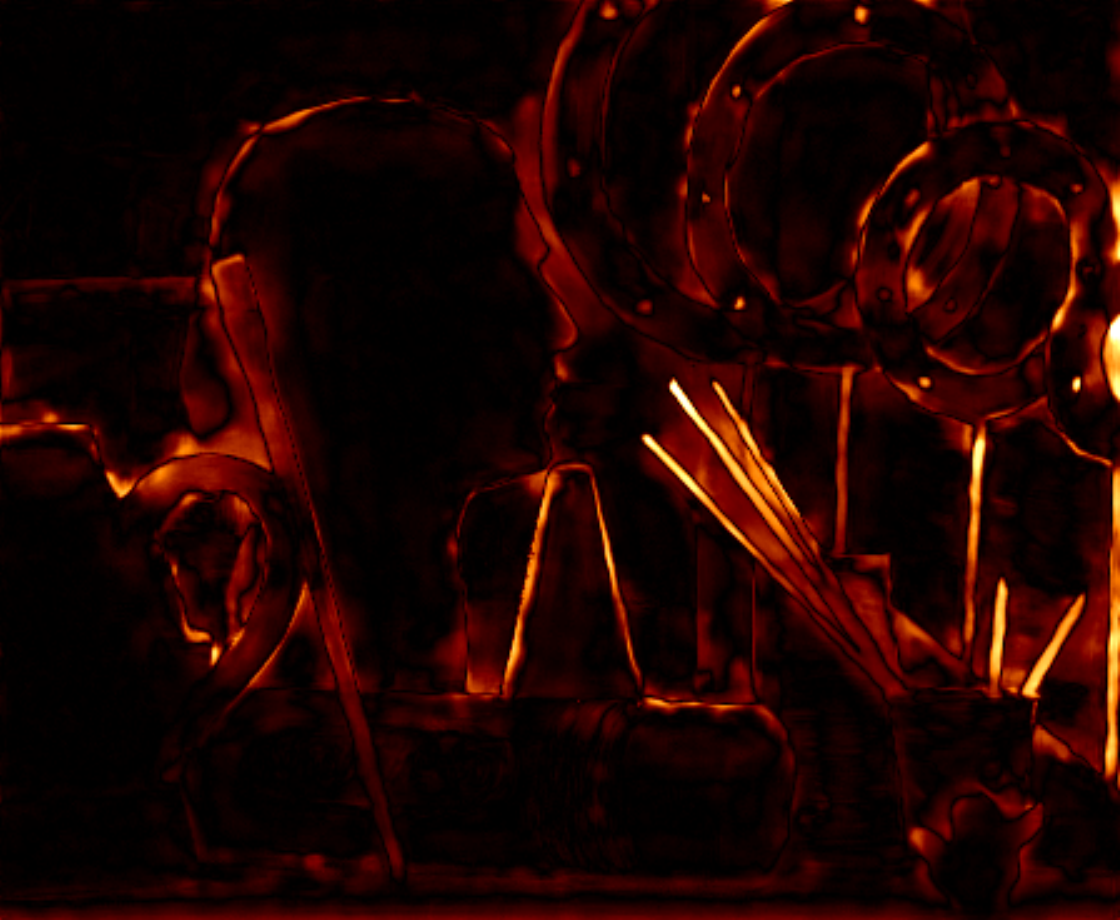}
			
			\vspace{\titlevspace}
			\caption*{(e) PAC~\cite{DBLP:conf/cvpr/SuJSGLK19}}
		\end{subfigure}	\hspace{\hhspace}
		
		\vspace{-0.1em}
		\begin{subfigure}{\widthh \linewidth}
			\centering
			\includegraphics[width=\lline \linewidth]{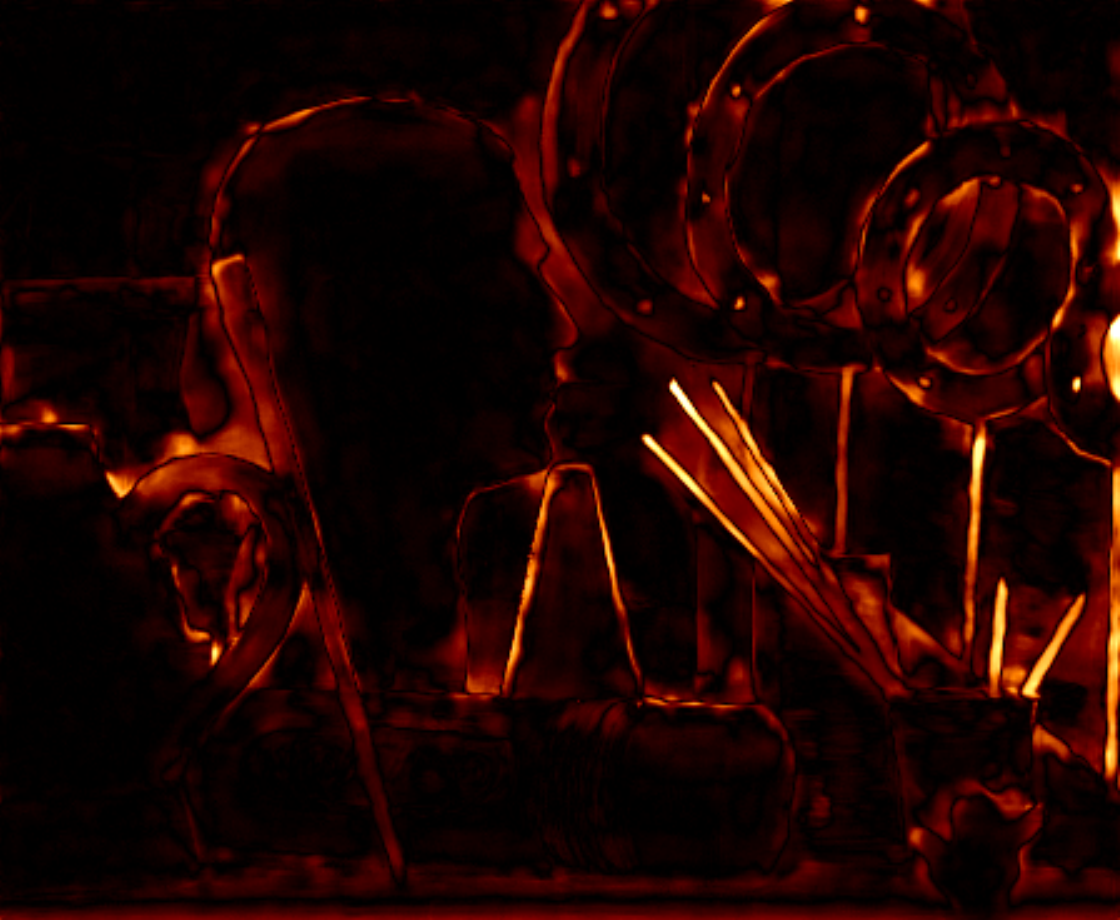}
			
			\vspace{\titlevspace}
			\caption*{(f) CUNet~\cite{DBLP:journals/pami/0002D21}}
		\end{subfigure}\hspace{\hhspace}
		\begin{subfigure}{\widthh \linewidth}
			\centering
			\includegraphics[width=\lline \linewidth]{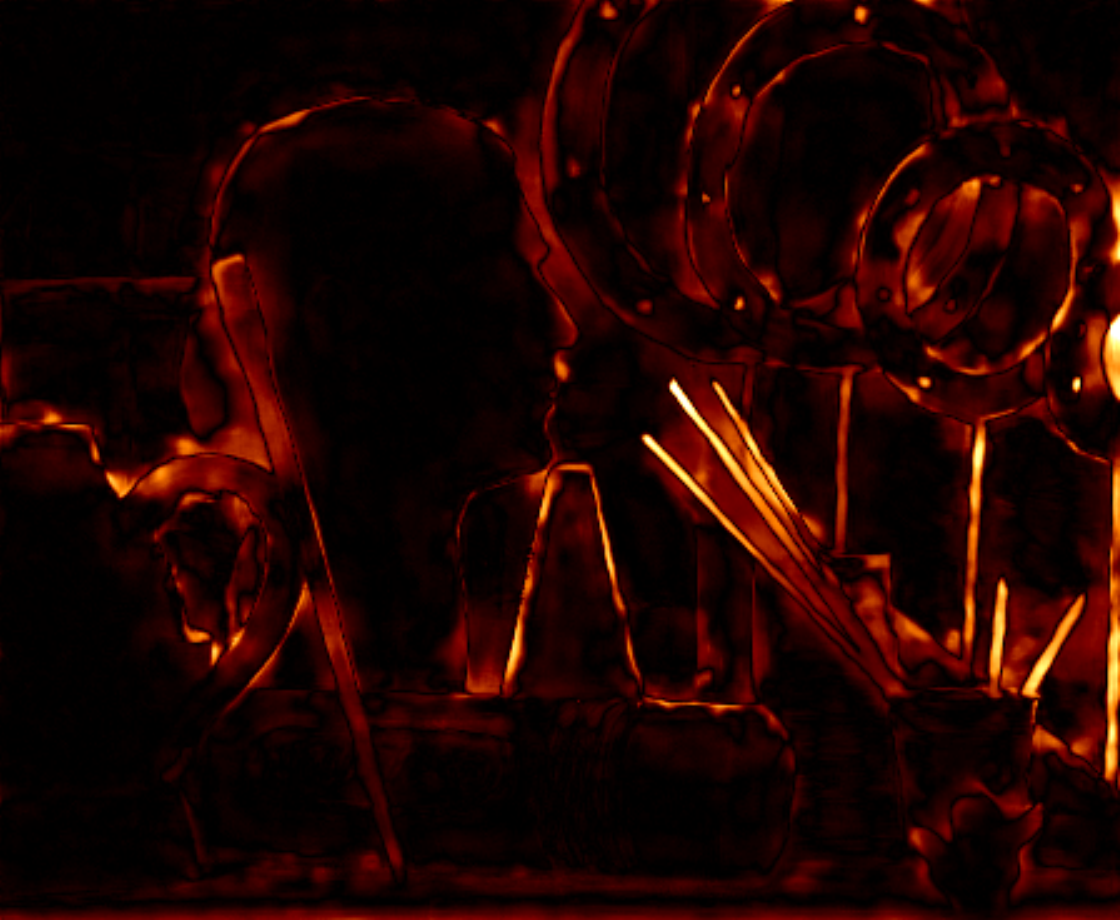}
			
			\vspace{\titlevspace}
			\caption*{(g) DKN~\cite{DBLP:journals/ijcv/KimPH21}}
		\end{subfigure}\hspace{\hhspace}
		\begin{subfigure}{\widthh \linewidth}
			\centering
			\includegraphics[width=\lline \linewidth]{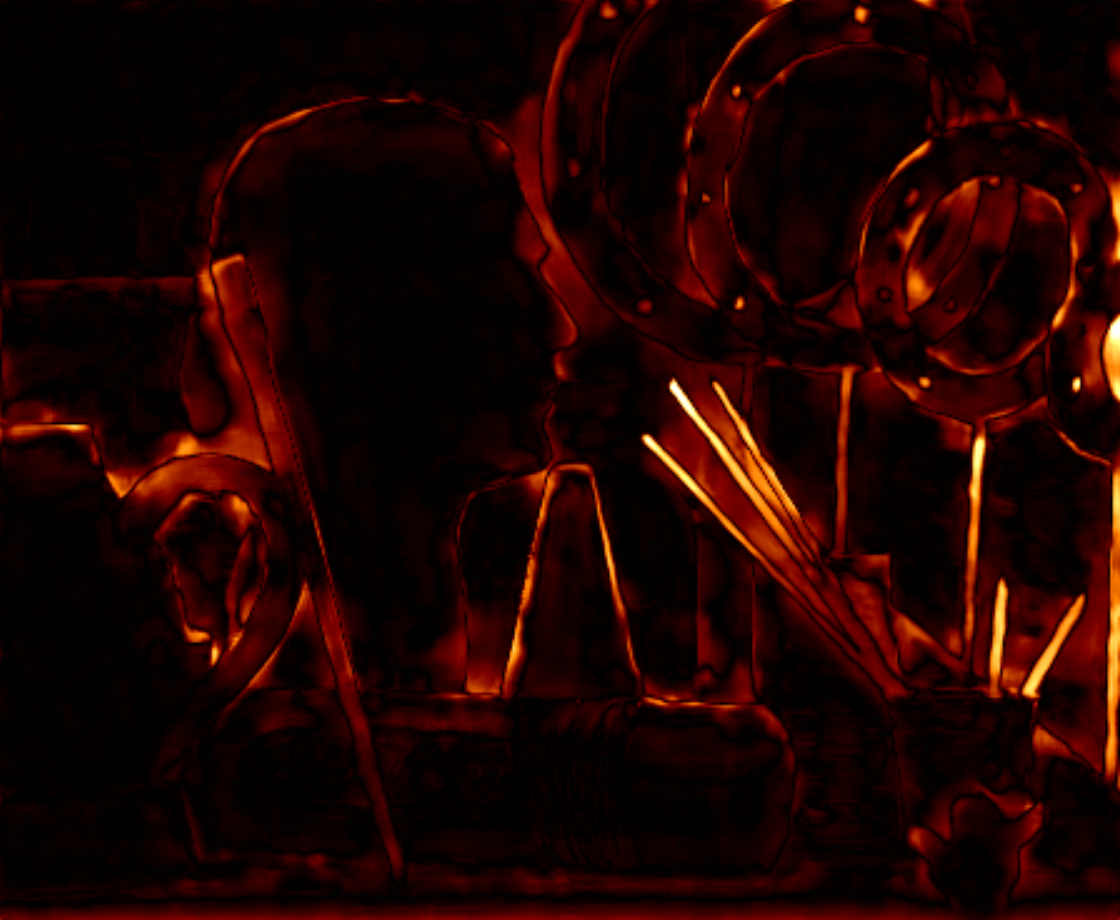}
			
			\vspace{\titlevspace}
			\caption*{(h) FDKN~\cite{DBLP:journals/ijcv/KimPH21}}
		\end{subfigure}\hspace{\hhspace}
		\begin{subfigure}{\widthh \linewidth}
			\centering
			\includegraphics[width=\lline \linewidth]{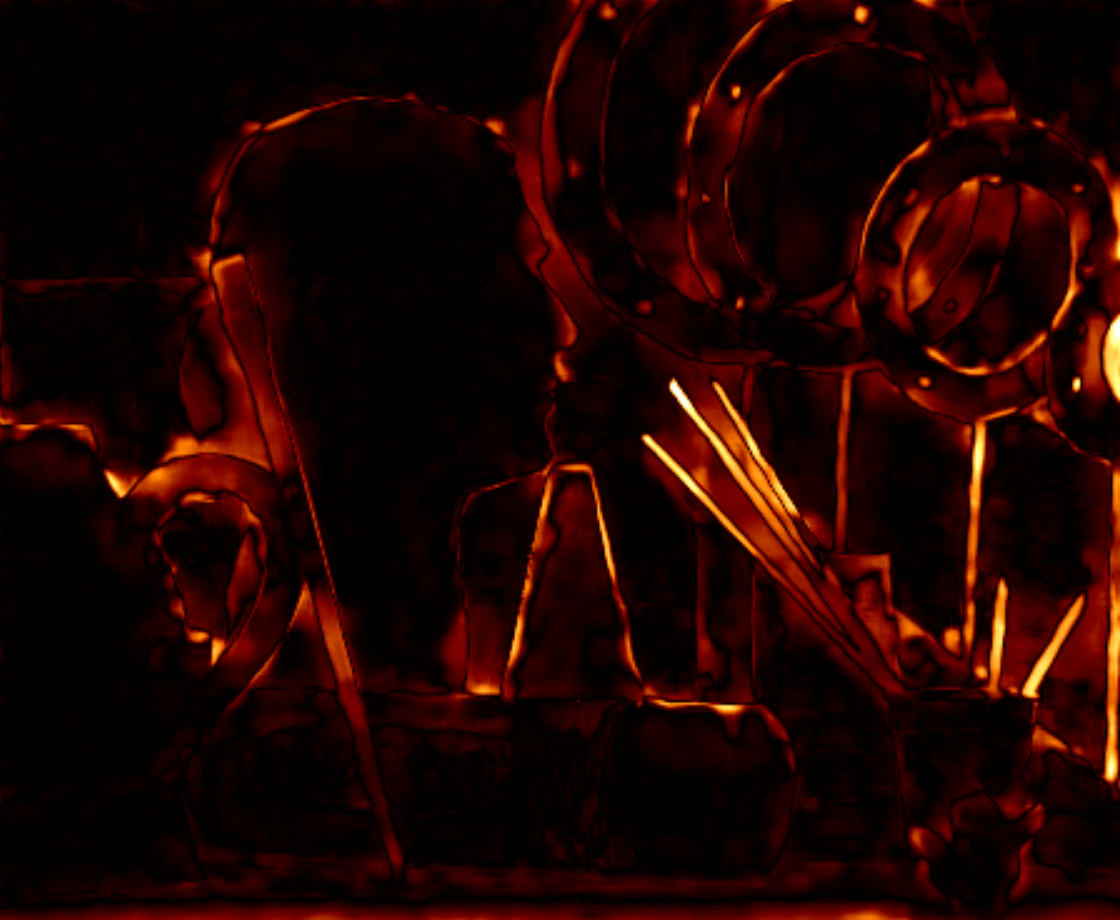}
			
			\vspace{\titlevspace}
			\caption*{(i) FDSR~\cite{DBLP:conf/cvpr/HeZLBCZLL021}}
		\end{subfigure}\hspace{\hhspace}
		\begin{subfigure}{\widthh \linewidth}
			\centering
			\includegraphics[width=\lline \linewidth]{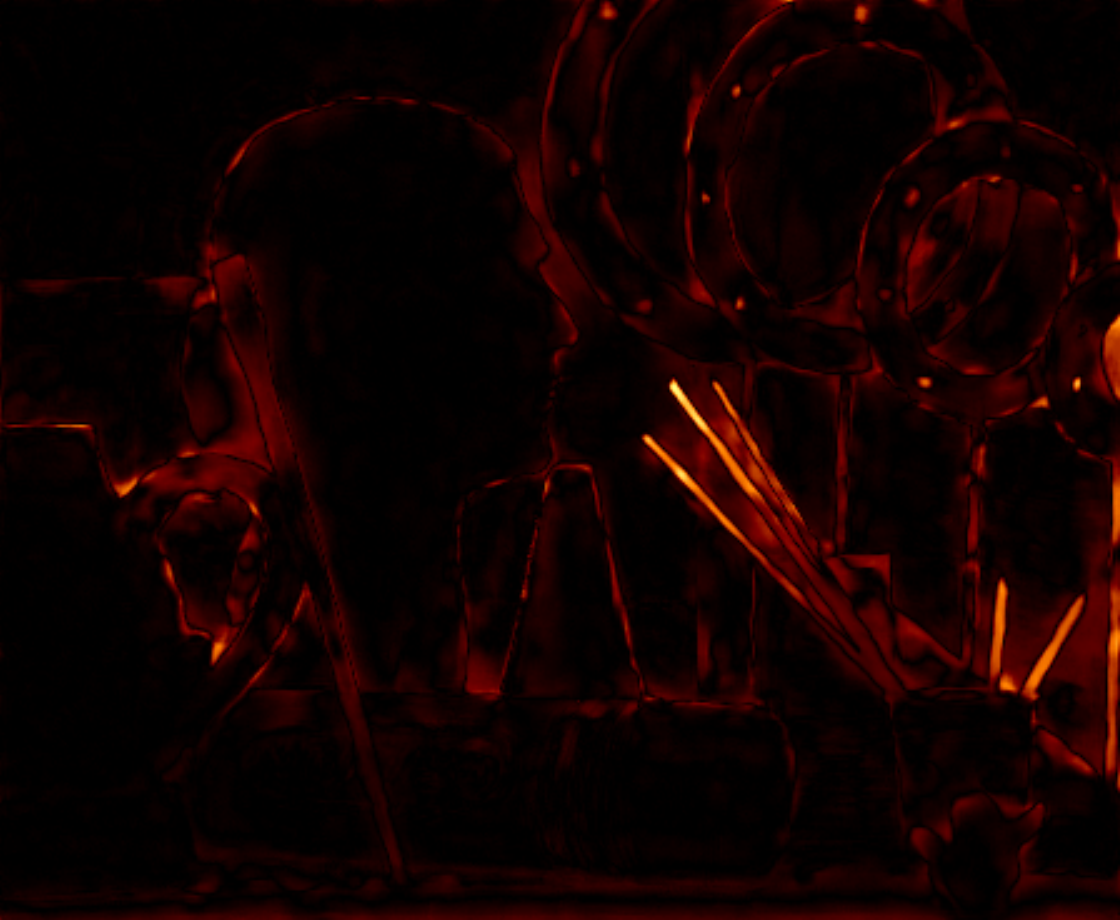}
			
			\vspace{\titlevspace}
			\caption*{(j) DCTNet ({\bf Ours})}
		\end{subfigure}	\hspace{\hhspace}
		
		\vspace{-1em}
		\caption{Visual comparison of error maps for ``05-Art'' in the Middlebury dataset for 16$\times$ super-resolution.}
		\label{fig:Qualitative2}
	\end{figure*}

	\begin{table*}[t]
		\centering
		\resizebox{0.77\linewidth}{!}{
			\begin{tabular}{lcccccccccccc}
				\toprule
				\multirow{2}{*}{Methods}                                                    &             \multicolumn{3}{c}{Middlebury}             &               \multicolumn{3}{c}{NYU V2}               &                 \multicolumn{3}{c}{Lu}                 &               \multicolumn{3}{c}{RGBDD}                \\
				\cmidrule(lr){2-4}\cmidrule(lr){5-7}\cmidrule(lr){8-10}\cmidrule(lr){11-13} &    $\times$4     &    $\times$8     &    $\times$16    &    $\times$4     &    $\times$8     &    $\times$16    &    $\times$4     &    $\times$8     &    $\times$16    &    $\times$4     &    $\times$8     &    $\times$16    \\ \midrule
				DJF~\cite{DBLP:conf/eccv/LiHA016}                                           &       1.68       &       3.24       &       5.62       &       2.80       &       5.33       &       9.46       &       1.65       &       3.96       &       6.75       &       3.41       &       5.57       &       8.15       \\
				DJFR~\cite{DBLP:journals/pami/LiHAY19}                                      &       1.32       &       3.19       &       5.57       &       2.38       &       4.94       &       9.18       &       1.15       &       3.57       &       6.77       &       3.35       &       5.57       &       7.99       \\
				PAC~\cite{DBLP:conf/cvpr/SuJSGLK19}                                         &       1.32       &       2.62       &       4.58       &       1.89       &       3.33       &       6.78       &       1.20       &       2.33       &       5.19       &       1.25       &       1.98       &       3.49       \\
				CUNet~\cite{DBLP:journals/pami/0002D21}                                     &       1.10       &       2.17       &       4.33       &       1.92       &       3.70       &       6.78       &       0.91       &       2.23       & \underline{4.99} &       1.18       &       1.95       &       3.45       \\
				DKN~\cite{DBLP:journals/ijcv/KimPH21}                                       &       1.23       &       2.12       & \underline{4.24} &       1.62       &       3.26       &       6.51       &       0.96       &       2.16       &       5.11       &       1.30       &       1.96       &       3.42       \\
				FDKN~\cite{DBLP:journals/ijcv/KimPH21}                                      &  \textbf{1.08}   &       2.17       &       4.50       &       1.86       &       3.58       &       6.96       &  \textbf{0.82}   & \underline{2.10} &       5.05       &       1.18       &       1.91       &       3.41       \\
				FDSR~\cite{DBLP:conf/cvpr/HeZLBCZLL021}                                     &       1.13       & \underline{2.08} &       4.39       & \underline{1.61} & \underline{3.18} & \underline{5.86} &       1.29       &       2.19       &       5.00       & \underline{1.16} & \underline{1.82} &  \underline{3.06}   \\
				DCTNet (Ours)                                                                        & \underline{1.10} &  \textbf{2.05}   &  \textbf{4.19}   &  \textbf{1.59}   &  \textbf{3.16}   &  \textbf{5.84}   & \underline{0.88} &  \textbf{1.85}   &  \textbf{4.39}   &  \textbf{1.08}   &  \textbf{1.74}   & \textbf{3.05} \\ \bottomrule
			\end{tabular}}
		
		\vspace{-1em}
		\caption{
        Quantitative comparison between our DCTNet and previous state-of-the-art approaches on four benchmark datasets. We use the RMSE metric (lower is better). The best and the second-best values are highlighted by \textbf{bold} and \underline{underline}, respectively. 
		}	
		\label{tab:quant}
	\end{table*}
	
	\begin{table}[ht]
		\centering
		\resizebox{0.7\linewidth}{!}{
			\begin{tabular}{lclc}
				\toprule
				Methods                                             & RMSE & Methods                                 &       RMSE       \\ \midrule
				SVLRM~\cite{DBLP:conf/cvpr/PanDRLT019}              & 8.05 & DKN~\cite{DBLP:journals/ijcv/KimPH21}   & \underline{7.38} \\
				DJF~\cite{DBLP:conf/eccv/LiHA016}                   & 7.90 & FDSR~\cite{DBLP:conf/cvpr/HeZLBCZLL021} &       7.50       \\
				DJFR~\cite{DBLP:journals/pami/LiHAY19}              & 8.01 & DCTNet                                  &  \textbf{7.37}   \\
				FDKN~\cite{DBLP:journals/ijcv/KimPH21}              & 7.50 &                                         &                  \\ \midrule
				FDSR$^*$~\cite{DBLP:conf/cvpr/HeZLBCZLL021}         & 5.49 & DCTNet$^*$                              &  \textbf{5.43}   \\ \bottomrule
			\end{tabular}}
        \vspace{-1em}
		\caption{Quantitative results on the \textit{real-world branch} of the RGBDD dataset. The best and second best values are highlighted by \textbf{bold} and \underline{underline}, respectively. FDSR$^*$ and DCTNet$^*$ represent the results after finetuning on real-world branch data.}	
		\label{tab:quant2}
	\end{table}
	
	\newcommand{\parameterimagewidth}{0.33}
	\begin{figure*}[t]
		\centering
		\begin{subfigure}{\parameterimagewidth\linewidth}
			\centering
			\includegraphics[width=\linewidth]{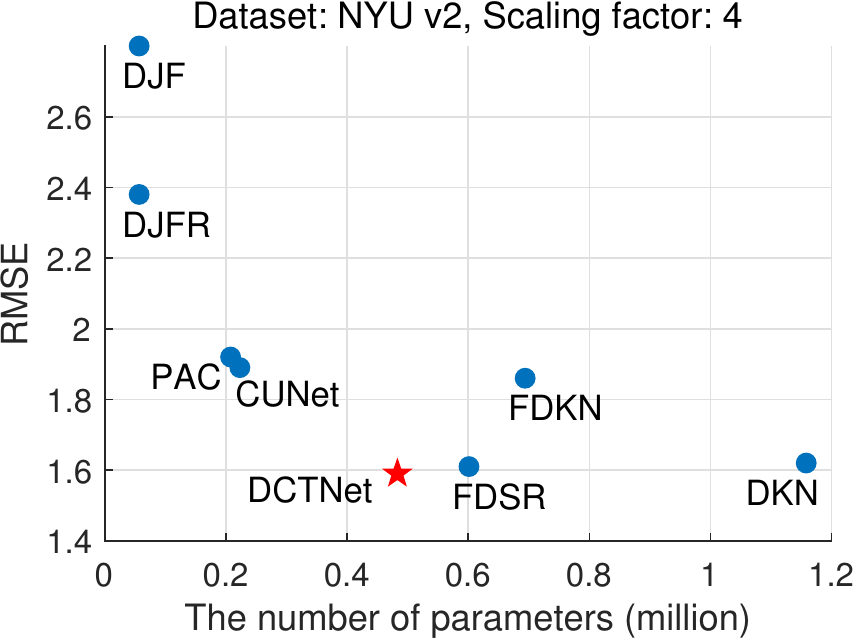}
		\end{subfigure}
		\begin{subfigure}{\parameterimagewidth\linewidth}
			\centering
			\includegraphics[width=\linewidth]{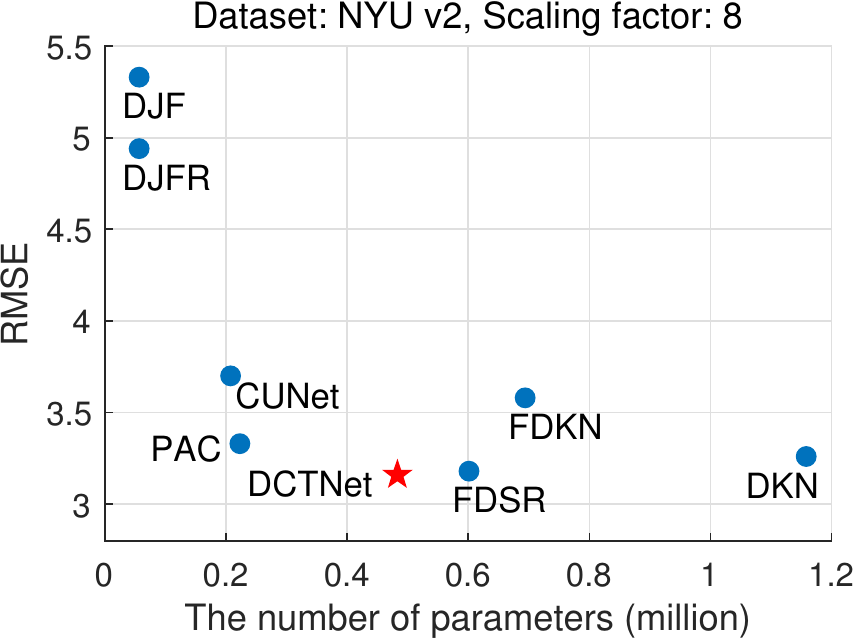}
		\end{subfigure}
		\begin{subfigure}{\parameterimagewidth\linewidth}
			\centering
			\includegraphics[width=0.97\linewidth]{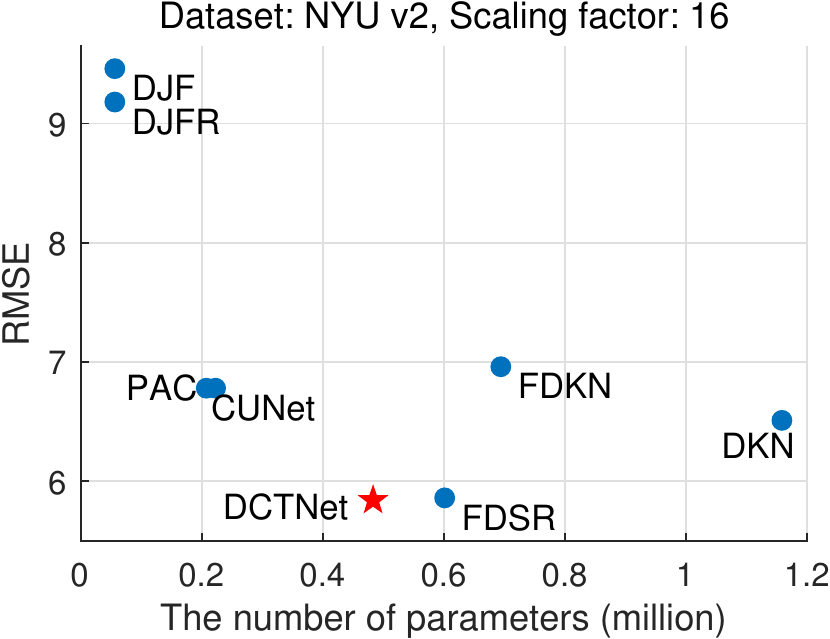}
		\end{subfigure}
		
		\vspace{-0.5em}
		\caption{
        The number of model parameters \vs RMSE on the NYU v2 dataset for $\times$4, $\times$8, and $\times$16 SR scales. Our DCTNet (red star) achieves better or comparable performance with a relatively small number of parameters than existing models (blue dots).
		}\label{fig:parameter}
	\end{figure*}
	
	\begin{table}[t]
		\centering
		\resizebox{0.8\linewidth}{!}{
			\begin{tabular}{ccccc}
				\toprule
				                                       &     {Configurations}     &   $\times$4   &   $\times$8   &  $\times$16   \\ \midrule
				\uppercase\expandafter{\romannumeral1} &  w/ Independent Filters  &     1.74      &     3.34      &     6.06      \\
				\uppercase\expandafter{\romannumeral2} & w/ Fully-shared Filters &     1.80      &     3.48      &     6.46      \\
				\uppercase\expandafter{\romannumeral3} &      w/o DCT Module      &     1.78      &     3.46      &     6.61      \\
				\uppercase\expandafter{\romannumeral4} & w/o Learnable Parameters &     1.77      &     3.55      &     6.63      \\
				\uppercase\expandafter{\romannumeral5} & w/o Residual Connection  &     1.91      &     3.84      &     7.06      \\ \midrule
				                                       &          {Ours}          & \textbf{1.59} & \textbf{3.16} & \textbf{5.84} \\ \bottomrule
			\end{tabular}}
			\vspace{-1em}
		\caption{Results of ablation experiments on the NYU v2 test set. \textbf{Bold} indicates the best score in terms of RMSE.}
		\label{tab:Ablation}
	\end{table}
	
	\subsection{Validation experiments}\label{sec:4_2}
	
	\bfsection{Impact of network depth and width}
    For our proposed DCTNet, the network depth $P$ and the width $C$ play an important role in the effectiveness of super-resolution. We show the results among different combinations of $\{P,C\}$ on the validation set. We first fix $C=64$, and calculate the prediction quality when $P=2,3,4,5,6$ on the validation set. Then we verify the SR results for $C=8,16,32,64,128$ when fixing $P=4$. The results are demonstrated in Tab.~\ref{tab:depth}. When $P<4$, the model capability is restricted. When $P>4$, increasing the depth does not achieve obvious performance gain but makes the model heavier. Similarly, when $C$ exceeds 64, there is no significant performance improvement but increases the training cost. To have a good balance of model performance and computational cost, we set $\{P=4, C=64\}$ for the following experiments.
	
	\bfsection{Highlighting edge attention weights} 
    We visualize the first three and last two channels of the guided edge attention weights $\tilde{\mathcal{W}}^R$ in Eq.~(\ref{eq:weight_attention}) from a representative sample pair (Fig.~\ref{fig:learnable_lambda}). We can clearly see that after the weight attention operation in the GESA module, the contour of the object is effectively highlighted, and the texture information inside the object is smoothed, which can alleviate the issue for texture over-transferred and benefit the GDSR task.
	
	\bfsection{Evolution of the learnable parameters in DCT}
    One of our contributions is to make the tuning parameter $\tilde\lambda$ in Eq.~(\ref{eq:DCT3}) a list of channel-wise learnable parameters to improve the flexibility of DCTNet. Here we show the changing curve of $\tilde\lambda$ in each channel against the iteration number during training  (\cref{fig:learnable_lambda}). The plot shows that under the data-driven setting, $\tilde\lambda$ can adaptively adjust the importance between the fidelity term and the regular term. Compared with the manually given $\lambda$ in Eq.~(\ref{eq:DCT2}), our design is more capable of leveraging the characteristics of different data domains.
	
	\subsection{Comparison with the state-of-the-arts}
	In this section, we test our DCTNet on the NYU v2, Middlebury, Lu and RGBDD benchmarks, and compare the results with state-of-the-art methods including DJF~\cite{DBLP:conf/eccv/LiHA016}, DJFR~\cite{DBLP:journals/pami/LiHAY19}, PAC~\cite{DBLP:conf/cvpr/SuJSGLK19}, CUNet~\cite{DBLP:journals/pami/0002D21}, DKN~\cite{DBLP:journals/ijcv/KimPH21}, FDKN~\cite{DBLP:journals/ijcv/KimPH21} and FDSR~\cite{DBLP:conf/cvpr/HeZLBCZLL021} to demonstrate its performance.
	
	\bfsection{Qualitative Comparison}
    We show the comparison of error maps for the SR depth maps in Fig.~\ref{fig:Qualitative1} and \ref{fig:Qualitative2}. Qualitatively, the depth predictions of DCTNet have lower prediction errors and are closer to the ground truth images. More visual comparisons are shown in the supplementary material.
	
	\bfsection{Quantitative Comparison}
    The quantitative results on four test sets with scaling factors $\times$4, $\times$8, and $\times$16 are shown in \cref{tab:quant}. Compared with existing approaches that only perform well on a certain dataset or super-resolution factor, our DCTNet achieves the best or second-best performance for multiple datasets and different super-resolution scales. This shows the advantage of our model upon previous state-of-the-arts. Moreover, following~\cite{DBLP:conf/cvpr/HeZLBCZLL021}, for the \textit{real-world branch} of RGBDD dataset, we use the $\times$4 models trained in Tab.~\ref{tab:quant} to verify their generalization ability in real-world scenes. All the models are tested directly without additional finetuning. The quantitative results are shown in Tab.~\ref{tab:quant2}. Our proposed DCTNet achieves lower RMSE than previous methods, demonstrating its generalization ability.
	
	\bfsection{Parameter Comparison}
    We discussed in Sec.~\ref{sec:3} that the semi-coupled feature extraction (SCFE) module and the DCT module can reduce the number of learnable parameters while improving the interpretability of the model. Therefore, we show the number of model parameters \vs RMSE on the NYU v2 dataset in Fig.~\ref{fig:parameter}. Our model compares favorably against existing approaches with a relatively small number of parameters, demonstrating promising future directions in building lightweight network architectures.
    
	
	\subsection{Ablation Studies}\label{sec:Ablation}
    We further validate the design choices of our DCTNet through ablation experiments (Tab.~\ref{tab:Ablation}). Due to space limitations, we refer readers to the supplementary material for details about the network structures in Exp.~\uppercase\expandafter{\romannumeral3} and \uppercase\expandafter{\romannumeral5}.
	
	\bfsection{Semi-coupled filters}
    Besides the default semi-coupled filters in the SCFE module, we also tested independent (Exp.~\uppercase\expandafter{\romannumeral1}) or fully-coupled (Exp.~\uppercase\expandafter{\romannumeral2}) cases, where the parameters in each residual block are not shared or fully-shared, respectively. Exp.~\uppercase\expandafter{\romannumeral1} result shows that the ability of independent kernels in extracting features is weaker than that of the semi-coupled filters, which demonstrates the necessity of using shared kernels to extract common features. On the other hand, Exp.~\uppercase\expandafter{\romannumeral2} shows fully-shared filters lead to worse performance than the semi-coupled ones, which indicates the importance of considering the disparity between two modalities. 
	
	\bfsection{The DCT module}
    In Exp.~\uppercase\expandafter{\romannumeral3}, we remove the DCT module and use a three-layer CNN to learn the mapping in Eq.~(\ref{eq:DCT4}). Removing the DCT module not only increases the number of learnable parameters but also reduces the prediction quality, which proves the effectiveness of the DCT module that follows the optimization-based methodology.
	
	\bfsection{Learnable parameters $\tilde\lambda$}
	Instead of using learnable ones, we fix $\tilde\lambda$ to $e^{0.1}$ in Exp.~\uppercase\expandafter{\romannumeral4} (the mean of their initialization values). The result shows that a fixed tuning coefficient can reduce the flexibility of the model and restrict the SR ability.
	
	\bfsection{Residual skip connection}
    In Exp.~\uppercase\expandafter{\romannumeral5}, we remove the residual connection in the SCFE module and only use a stack of convolution kernels. The results show that residual connections play an important role in the feature extraction stage, as removing them leads to significant performance degradation.
	
	\subsection{Limitation}\label{sec:Limitation}
    Although our proposed DCTNet is more interpretable and compares favorably against existing approaches with less learnable parameters, one limitation is that the components in the model make the formulation more complex than approaches using an end-to-end deep neural network to regress the HR depth map. In our future work, we will explore different ways to simplify the network design while keeping the merit of interpretability and the good tradeoff between network parameters and SR performance. We will also investigate challenges ubiquitous for most guided depth SR approaches, \eg, low illumination and blurry boundary in the paired RGB images.
	
	\section{Conclusion}\label{sec:5}
    In this paper, we propose a novel guided depth super-resolution (GDSR) model, DCTNet, based on discrete cosine transform, semi-coupled convolutional feature extraction, and adaptive edge attention. Our DCTNet incorporates intuitive motivations into the design choices to alleviate the challenges of RGB texture over-transferred, ineffective cross-modal feature extraction, and unclear working mechanism of network components in existing methods. In the future, we hope that more multi-modal image processing tasks can benefit from all or some components in DCTNet.
	
	\section*{Acknowledgement}\label{sec:6}
	This work has been supported by the National Natural Science Foundation of China under Grant 61976174.

	{
	\small
	\bibliographystyle{ieee_fullname}
	\bibliography{main}

\begin{thebibliography}{10}\itemsep=-1pt

\bibitem{DBLP:journals/tcyb/CamplaniMS13}
Massimo Camplani, Tom{\'{a}}s Mantec{\'{o}}n, and Luis Salgado.
\newblock Depth-color fusion strategy for 3-d scene modeling with kinect.
\newblock {\em {IEEE} Trans. Cybern.}, 43(6):1560--1571, 2013.

\bibitem{DBLP:journals/tip/DengD20}
Xin Deng and Pier~Luigi Dragotti.
\newblock Deep coupled {ISTA} network for multi-modal image super-resolution.
\newblock {\em {IEEE} Trans. Image Process.}, 29:1683--1698, 2020.

\bibitem{DBLP:journals/pami/0002D21}
Xin Deng and Pier~Luigi Dragotti.
\newblock Deep convolutional neural network for multi-modal image restoration
  and fusion.
\newblock {\em {IEEE} Trans. Pattern Anal. Mach. Intell.}, 43(10):3333--3348,
  2021.

\bibitem{DBLP:conf/nips/DiebelT05}
James Diebel and Sebastian Thrun.
\newblock An application of markov random fields to range sensing.
\newblock In {\em {NIPS}}, pages 291--298, 2005.

\bibitem{DBLP:conf/eccv/DongLHT14}
Chao Dong, Chen~Change Loy, Kaiming He, and Xiaoou Tang.
\newblock Learning a deep convolutional network for image super-resolution.
\newblock In {\em {ECCV}}, pages 184--199. Springer, 2014.

\bibitem{DBLP:conf/iccv/FerstlRRRB13}
David Ferstl, Christian Reinbacher, Ren{\'{e}} Ranftl, Matthias R{\"{u}}ther,
  and Horst Bischof.
\newblock Image guided depth upsampling using anisotropic total generalized
  variation.
\newblock In {\em {ICCV}}, pages 993--1000. {IEEE}, 2013.

\bibitem{DBLP:conf/icml/GregorL10}
Karol Gregor and Yann LeCun.
\newblock Learning fast approximations of sparse coding.
\newblock In {\em {ICML}}, pages 399--406. Omnipress, 2010.

\bibitem{DBLP:conf/cvpr/GuZGCCZ17}
Shuhang Gu, Wangmeng Zuo, Shi Guo, Yunjin Chen, Chongyu Chen, and Lei Zhang.
\newblock Learning dynamic guidance for depth image enhancement.
\newblock In {\em {CVPR}}, pages 712--721. {IEEE} Computer Society, 2017.

\bibitem{DBLP:journals/tip/GuoLGCFH19}
Chunle Guo, Chongyi Li, Jichang Guo, Runmin Cong, Huazhu Fu, and Ping Han.
\newblock Hierarchical features driven residual learning for depth map
  super-resolution.
\newblock {\em {IEEE} Trans. Image Process.}, 28(5):2545--2557, 2019.

\bibitem{DBLP:conf/eccv/GuptaGAM14}
Saurabh Gupta, Ross~B. Girshick, Pablo~Andr{\'{e}}s Arbel{\'{a}}ez, and
  Jitendra Malik.
\newblock Learning rich features from {RGB-D} images for object detection and
  segmentation.
\newblock In {\em {ECCV}}, pages 345--360. Springer, 2014.

\bibitem{DBLP:journals/pami/He0T13}
Kaiming He, Jian Sun, and Xiaoou Tang.
\newblock Guided image filtering.
\newblock {\em {IEEE} Trans. Pattern Anal. Mach. Intell.}, 35(6):1397--1409,
  2013.

\bibitem{DBLP:conf/cvpr/HeZLBCZLL021}
Lingzhi He, Hongguang Zhu, Feng Li, Huihui Bai, Runmin Cong, Chunjie Zhang,
  Chunyu Lin, Meiqin Liu, and Yao Zhao.
\newblock Towards fast and accurate real-world depth super-resolution:
  Benchmark dataset and baseline.
\newblock In {\em {CVPR}}, pages 9229--9238. {IEEE} Computer Society, 2021.

\bibitem{DBLP:conf/cvpr/HirschmullerS07}
Heiko Hirschm{\"{u}}ller and Daniel Scharstein.
\newblock Evaluation of cost functions for stereo matching.
\newblock In {\em {CVPR}}. {IEEE} Computer Society, 2007.

\bibitem{DBLP:conf/eccv/HuiLT16}
Tak{-}Wai Hui, Chen~Change Loy, and Xiaoou Tang.
\newblock Depth map super-resolution by deep multi-scale guidance.
\newblock In {\em {ECCV}}, pages 353--369. Springer, 2016.

\bibitem{DBLP:conf/iccv/KiechleHK13}
Martin Kiechle, Simon Hawe, and Martin Kleinsteuber.
\newblock A joint intensity and depth co-sparse analysis model for depth map
  super-resolution.
\newblock In {\em {ICCV}}, pages 1545--1552. {IEEE}, 2013.

\bibitem{DBLP:journals/ijcv/KimPH21}
Beomjun Kim, Jean Ponce, and Bumsub Ham.
\newblock Deformable kernel networks for joint image filtering.
\newblock {\em Int. J. Comput. Vis.}, 129(2):579--600, 2021.

\bibitem{kingma2014adam}
Diederik Kingma and Jimmy Ba.
\newblock Adam: A method for stochastic optimization.
\newblock In {\em ICLR}, 2014.

\bibitem{DBLP:journals/tog/KopfCLU07}
Johannes Kopf, Michael~F. Cohen, Dani Lischinski, and Matthew Uyttendaele.
\newblock Joint bilateral upsampling.
\newblock {\em {ACM} Trans. Graph.}, 26(3):96, 2007.

\bibitem{DBLP:conf/cvpr/KwonTL15}
HyeokHyen Kwon, Yu{-}Wing Tai, and Stephen Lin.
\newblock Data-driven depth map refinement via multi-scale sparse
  representation.
\newblock In {\em {CVPR}}, pages 159--167. {IEEE} Computer Society, 2015.

\bibitem{DBLP:journals/tii/LiLY0J021}
Ling Li, Xiaojian Li, Shanlin Yang, Shuai Ding, Alireza Jolfaei, and Xi Zheng.
\newblock Unsupervised-learning-based continuous depth and motion estimation
  with monocular endoscopy for virtual reality minimally invasive surgery.
\newblock {\em {IEEE} Trans. Ind. Informatics}, 17(6):3920--3928, 2021.

\bibitem{DBLP:conf/eccv/LiHA016}
Yijun Li, Jia{-}Bin Huang, Narendra Ahuja, and Ming{-}Hsuan Yang.
\newblock Deep joint image filtering.
\newblock In {\em {ECCV}}, pages 154--169. Springer, 2016.

\bibitem{DBLP:journals/pami/LiHAY19}
Yijun Li, Jia{-}Bin Huang, Narendra Ahuja, and Ming{-}Hsuan Yang.
\newblock Joint image filtering with deep convolutional networks.
\newblock {\em {IEEE} Trans. Pattern Anal. Mach. Intell.}, 41(8):1909--1923,
  2019.

\bibitem{DBLP:conf/eccv/LiMDL16}
Yu Li, Dongbo Min, Minh~N. Do, and Jiangbo Lu.
\newblock Fast guided global interpolation for depth and motion.
\newblock In {\em {ECCV}}, pages 717--733. Springer, 2016.

\bibitem{DBLP:conf/eccv/LiaoLZZLY20}
Miao Liao, Feixiang Lu, Dingfu Zhou, Sibo Zhang, Wei Li, and Ruigang Yang.
\newblock {DVI:} depth guided video inpainting for autonomous driving.
\newblock In {\em {ECCV}}, pages 1--17. Springer, 2020.

\bibitem{lim2017enhanced}
Bee Lim, Sanghyun Son, Heewon Kim, Seungjun Nah, and Kyoung~Mu Lee.
\newblock Enhanced deep residual networks for single image super-resolution.
\newblock In {\em CVPRW}, 2017.

\bibitem{DBLP:conf/cvpr/Liu0T0W20}
Jie Liu, Wenjie Zhang, Yuting Tang, Jie Tang, and Gangshan Wu.
\newblock Residual feature aggregation network for image super-resolution.
\newblock In {\em {CVPR}}, pages 2356--2365. {IEEE} Computer Society, 2020.

\bibitem{DBLP:conf/cvpr/0001TT13}
Ming{-}Yu Liu, Oncel Tuzel, and Yuichi Taguchi.
\newblock Joint geodesic upsampling of depth images.
\newblock In {\em {CVPR}}, pages 169--176. {IEEE} Computer Society, 2013.

\bibitem{DBLP:journals/tip/LiuZCJZG19a}
Xianming Liu, Deming Zhai, Rong Chen, Xiangyang Ji, Debin Zhao, and Wen Gao.
\newblock Depth super-resolution via joint color-guided internal and external
  regularizations.
\newblock {\em {IEEE} Trans. Image Process.}, 28(4):1636--1645, 2019.

\bibitem{DBLP:conf/cvpr/LuF15}
Jiajun Lu and David~A. Forsyth.
\newblock Sparse depth super resolution.
\newblock In {\em {CVPR}}, pages 2245--2253. {IEEE} Computer Society, 2015.

\bibitem{DBLP:conf/cvpr/LuSMLD12}
Jiangbo Lu, Keyang Shi, Dongbo Min, Liang Lin, and Minh~N. Do.
\newblock Cross-based local multipoint filtering.
\newblock In {\em {CVPR}}, pages 430--437. {IEEE} Computer Society, 2012.

\bibitem{DBLP:conf/cvpr/LuRL14}
Si Lu, Xiaofeng Ren, and Feng Liu.
\newblock Depth enhancement via low-rank matrix completion.
\newblock In {\em {CVPR}}, pages 3390--3397. {IEEE} Computer Society, 2014.

\bibitem{magid2021dynamic}
Salma~Abdel Magid, Yulun Zhang, Donglai Wei, Won{-}Dong Jang, Zudi Lin, Yun Fu,
  and Hanspeter Pfister.
\newblock Dynamic high-pass filtering and multi-spectral attention for image
  super-resolution.
\newblock In {\em {ICCV}}, pages 4268--4277. {IEEE}, 2021.

\bibitem{DBLP:journals/tip/MinLD12}
Dongbo Min, Jiangbo Lu, and Minh~N. Do.
\newblock Depth video enhancement based on weighted mode filtering.
\newblock {\em {IEEE} Trans. Image Process.}, 21(3):1176--1190, 2012.

\bibitem{DBLP:conf/cvpr/PanDRLT019}
Jinshan Pan, Jiangxin Dong, Jimmy S.~J. Ren, Liang Lin, Jinhui Tang, and
  Ming{-}Hsuan Yang.
\newblock Spatially variant linear representation models for joint filtering.
\newblock In {\em {CVPR}}, pages 1702--1711. {IEEE} Computer Society, 2019.

\bibitem{DBLP:conf/iccv/ParkKTBK11}
Jaesik Park, Hyeongwoo Kim, Yu{-}Wing Tai, Michael~S. Brown, and In{-}So Kweon.
\newblock High quality depth map upsampling for 3d-tof cameras.
\newblock In {\em {ICCV}}, pages 1623--1630. {IEEE}, 2011.

\bibitem{NEURIPS2019_9015}
Adam Paszke, Sam Gross, Francisco Massa, Adam Lerer, James Bradbury, Gregory
  Chanan, Trevor Killeen, Zeming Lin, Natalia Gimelshein, Luca Antiga, Alban
  Desmaison, Andreas Kopf, Edward Yang, Zachary DeVito, Martin Raison, Alykhan
  Tejani, Sasank Chilamkurthy, Benoit Steiner, Lu Fang, Junjie Bai, and Soumith
  Chintala.
\newblock Pytorch: An imperative style, high-performance deep learning library.
\newblock In H. Wallach, H. Larochelle, A. Beygelzimer, F. d\textquotesingle
  Alch\'{e}-Buc, E. Fox, and R. Garnett, editors, {\em Advances in Neural
  Information Processing Systems 32}, pages 8024--8035. Curran Associates,
  Inc., 2019.

\bibitem{DBLP:conf/cvpr/PengPLS20}
Wanli Peng, Hao Pan, He Liu, and Yi Sun.
\newblock {IDA-3D:} instance-depth-aware 3d object detection from stereo vision
  for autonomous driving.
\newblock In {\em {CVPR}}, pages 13012--13021. {IEEE} Computer Society, 2020.

\bibitem{DBLP:journals/corr/abs-2109-00212}
Haotong Qin, Yifu Ding, Xiangguo Zhang, Aoyu Li, Jiakai Wang, Xianglong Liu,
  and Jiwen Lu.
\newblock Diverse sample generation: Pushing the limit of data-free
  quantization.
\newblock {\em CoRR}, abs/2109.00212, 2021.

\bibitem{DBLP:conf/bmvc/RieglerFRB16}
Gernot Riegler, David Ferstl, Matthias R{\"{u}}ther, and Horst Bischof.
\newblock A deep primal-dual network for guided depth super-resolution.
\newblock In {\em {BMVC}}. {BMVA} Press, 2016.

\bibitem{DBLP:conf/eccv/RieglerRB16}
Gernot Riegler, Matthias R{\"{u}}ther, and Horst Bischof.
\newblock Atgv-net: Accurate depth super-resolution.
\newblock In {\em {ECCV}}, pages 268--284. Springer, 2016.

\bibitem{DBLP:conf/cvpr/ScharsteinP07}
Daniel Scharstein and Chris Pal.
\newblock Learning conditional random fields for stereo.
\newblock In {\em {CVPR}}. {IEEE} Computer Society, 2007.

\bibitem{DBLP:journals/pami/ShottonGFSCFMKCKB13}
Jamie Shotton, Ross~B. Girshick, Andrew~W. Fitzgibbon, Toby Sharp, Mat Cook,
  Mark Finocchio, Richard Moore, Pushmeet Kohli, Antonio Criminisi, Alex
  Kipman, and Andrew Blake.
\newblock Efficient human pose estimation from single depth images.
\newblock {\em {IEEE} Trans. Pattern Anal. Mach. Intell.}, 35(12):2821--2840,
  2013.

\bibitem{DBLP:conf/eccv/SilbermanHKF12}
Nathan Silberman, Derek Hoiem, Pushmeet Kohli, and Rob Fergus.
\newblock Indoor segmentation and support inference from {RGBD} images.
\newblock In {\em {ECCV}}, pages 746--760. Springer, 2012.

\bibitem{DBLP:conf/cvpr/SongDZLLLY20}
Xibin Song, Yuchao Dai, Dingfu Zhou, Liu Liu, Wei Li, Hongdong Li, and Ruigang
  Yang.
\newblock Channel attention based iterative residual learning for depth map
  super-resolution.
\newblock In {\em {CVPR}}, pages 5630--5639. {IEEE} Computer Society, 2020.

\bibitem{strang1999discrete}
Gilbert Strang.
\newblock The discrete cosine transform.
\newblock {\em SIAM review}, 41(1):135--147, 1999.

\bibitem{DBLP:conf/cvpr/SuJSGLK19}
Hang Su, Varun Jampani, Deqing Sun, Orazio Gallo, Erik~G. Learned{-}Miller, and
  Jan Kautz.
\newblock Pixel-adaptive convolutional neural networks.
\newblock In {\em {CVPR}}, pages 11166--11175. {IEEE} Computer Society, 2019.

\bibitem{DBLP:conf/cvpr/SunYLLW021}
Baoli Sun, Xinchen Ye, Baopu Li, Haojie Li, Zhihui Wang, and Rui Xu.
\newblock Learning scene structure guidance via cross-task knowledge transfer
  for single depth super-resolution.
\newblock In {\em {CVPR}}, pages 7792--7801. {IEEE} Computer Society, 2021.

\bibitem{DBLP:conf/cvpr/TanSP14}
Xiao Tan, Changming Sun, and Tuan~D. Pham.
\newblock Multipoint filtering with local polynomial approximation and range
  guidance.
\newblock In {\em {CVPR}}, pages 2941--2948. {IEEE} Computer Society, 2014.

\bibitem{DBLP:conf/mm/TangCZ21}
Jiaxiang Tang, Xiaokang Chen, and Gang Zeng.
\newblock Joint implicit image function for guided depth super-resolution.
\newblock In {\em {ACM} Multimedia}, pages 4390--4399. {ACM}, 2021.

\bibitem{DBLP:conf/mm/TangCSHZZK21}
Qi Tang, Runmin Cong, Ronghui Sheng, Lingzhi He, Dan Zhang, Yao Zhao, and Sam
  Kwong.
\newblock Bridgenet: {A} joint learning network of depth map super-resolution
  and monocular depth estimation.
\newblock In {\em {ACM} Multimedia}, pages 2148--2157. {ACM}, 2021.

\bibitem{DBLP:journals/tip/TosicD14}
Ivana Tosic and Sarah Drewes.
\newblock Learning joint intensity-depth sparse representations.
\newblock {\em {IEEE} Trans. Image Process.}, 23(5):2122--2132, 2014.

\bibitem{DBLP:journals/tip/WenSLLF19}
Yang Wen, Bin Sheng, Ping Li, Weiyao Lin, and David~Dagan Feng.
\newblock Deep color guided coarse-to-fine convolutional network cascade for
  depth image super-resolution.
\newblock {\em {IEEE} Trans. Image Process.}, 28(2):994--1006, 2019.

\bibitem{DBLP:conf/cvpr/Wu0ZH18}
Huikai Wu, Shuai Zheng, Junge Zhang, and Kaiqi Huang.
\newblock Fast end-to-end trainable guided filter.
\newblock In {\em {CVPR}}, pages 1838--1847. {IEEE} Computer Society, 2018.

\bibitem{DBLP:journals/tip/XieFS16}
Jun Xie, Rog{\'{e}}rio~Schmidt Feris, and Ming{-}Ting Sun.
\newblock Edge-guided single depth image super resolution.
\newblock {\em {IEEE} Trans. Image Process.}, 25(1):428--438, 2016.

\bibitem{DBLP:journals/tmm/XieFYS15}
Jun Xie, Rog{\'{e}}rio~Schmidt Feris, Shiaw{-}Shian Yu, and Ming{-}Ting Sun.
\newblock Joint super resolution and denoising from a single depth image.
\newblock {\em {IEEE} Trans. Multim.}, 17(9):1525--1537, 2015.

\bibitem{DBLP:conf/iccv/XiongZ0CYZY19}
Fu Xiong, Boshen Zhang, Yang Xiao, Zhiguo Cao, Taidong Yu, Joey~Tianyi Zhou,
  and Junsong Yuan.
\newblock {A2J:} anchor-to-joint regression network for 3d articulated pose
  estimation from a single depth image.
\newblock In {\em {ICCV}}, pages 793--802. {IEEE}, 2019.

\bibitem{xu2020learning}
Kai Xu, Minghai Qin, Fei Sun, Yuhao Wang, Yen-Kuang Chen, and Fengbo Ren.
\newblock Learning in the frequency domain.
\newblock In {\em {CVPR}}, pages 1740--1749. {IEEE} Computer Society, 2020.

\bibitem{DBLP:conf/cvpr/Xu0ZSL021}
Shuang Xu, Jiangshe Zhang, Zixiang Zhao, Kai Sun, Junmin Liu, and Chunxia
  Zhang.
\newblock Deep gradient projection networks for pan-sharpening.
\newblock In {\em {CVPR}}, pages 1366--1375. Computer Vision Foundation /
  {IEEE}, 2021.

\bibitem{DBLP:journals/tip/YangYLHW14}
Jingyu Yang, Xinchen Ye, Kun Li, Chunping Hou, and Yao Wang.
\newblock Color-guided depth recovery from {RGB-D} data using an adaptive
  autoregressive model.
\newblock {\em {IEEE} Trans. Image Process.}, 23(8):3443--3458, 2014.

\bibitem{DBLP:conf/cvpr/YangYDN07}
Qingxiong Yang, Ruigang Yang, James Davis, and David Nist{\'{e}}r.
\newblock Spatial-depth super resolution for range images.
\newblock In {\em {CVPR}}. {IEEE} Computer Society, 2007.

\bibitem{DBLP:journals/tip/YeSWYXLL20}
Xinchen Ye, Baoli Sun, Zhihui Wang, Jingyu Yang, Rui Xu, Haojie Li, and Baopu
  Li.
\newblock Pmbanet: Progressive multi-branch aggregation network for scene depth
  super-resolution.
\newblock {\em {IEEE} Trans. Image Process.}, 29:7427--7442, 2020.

\bibitem{DBLP:conf/cvpr/ZhangGT20}
Kai Zhang, Luc~Van Gool, and Radu Timofte.
\newblock Deep unfolding network for image super-resolution.
\newblock In {\em {CVPR}}, pages 3214--3223. {IEEE} Computer Society, 2020.

\bibitem{DBLP:conf/cvpr/ZhangZGZ17}
Kai Zhang, Wangmeng Zuo, Shuhang Gu, and Lei Zhang.
\newblock Learning deep {CNN} denoiser prior for image restoration.
\newblock In {\em {CVPR}}, pages 2808--2817. {IEEE} Computer Society, 2017.

\bibitem{DBLP:conf/iccv/ZhangBKIX17}
Yinda Zhang, Mingru Bai, Pushmeet Kohli, Shahram Izadi, and Jianxiong Xiao.
\newblock Deepcontext: Context-encoding neural pathways for 3d holistic scene
  understanding.
\newblock In {\em {ICCV}}, pages 1201--1210. {IEEE}, 2017.

\end{thebibliography}
	}
	
\end{document}